\documentclass[letterpaper]{article} 
\usepackage{aaai2026}  
\usepackage{times}  
\usepackage{helvet}  
\usepackage{courier}  
\usepackage[hyphens]{url}  
\usepackage{graphicx} 
\usepackage{natbib}  
\usepackage{caption} 
\usepackage{algorithm}
\usepackage{algorithmic}
\usepackage{amsmath}
\usepackage{amssymb}
\usepackage{multirow}
\usepackage{booktabs}
\usepackage{xcolor}

\usepackage{newfloat}
\usepackage{listings}
\DeclareCaptionStyle{ruled}{labelfont=normalfont,labelsep=colon,strut=off} 
\floatstyle{ruled}
\newfloat{listing}{tb}{lst}{}
\floatname{listing}{Listing}

\title{Nexus-Gen: Unified Image Understanding, Generation, and Editing via Prefilled Autoregression in Shared Embedding Space}

\author {
Hong Zhang$^{1,2}$,
Zhongjie Duan$^{2}$,
Xingjun Wang$^2$,
Yuze Zhao$^2$, \\
Weiyi Lu $^3$,
Zhipeng Di $^3$,
Yixuan Xu $^3$,
Yingda Chen$^2$,
Yu Zhang$^{1}$\footnotemark[2] \\
}
\affiliations {
    \textsuperscript{\rm 1}College of Control Science and Engineering, Zhejiang University \\
    \textsuperscript{\rm 2}ModelScope Team, Alibaba Group Inc.
    \textsuperscript{\rm 3}AIOS Team, Alibaba Group Inc. \\
\{hongzhang99, zhangyu80\}@zju.edu.cn \\
\{duanzhongjie.dzj, xingjun.wxj, yuze.zyz, weiyi.lwy, dizhipeng.dzp, xuyixuan.xyx, yingda.chen \}@alibaba-inc.com
}

\usepackage{bibentry}

\begin{document}
\nocopyright
\maketitle

\begin{abstract}
Unified multimodal generative models aim to integrate image understanding and generation abilities, offering significant advantages in harnessing multimodal corpora, particularly interleaved text-image data. However, existing unified models exhibit limitations in image synthesis quality, autoregressive error accumulation, and image editing capability. In this work, we propose Nexus-Gen, a novel architecture that unifies image understanding, generation, and editing tasks in a shared image embedding space. This shared space serves as a bridge for the autoregressive and diffusion models, which seamlessly integrates their complementary strengths in cross-modal modeling. To mitigate the severe error accumulation during autoregressive embedding prediction, we propose a novel prefilled autoregression strategy that aligns training-inference dynamics by prefilling input sequences with learnable embeddings. After multi-stage and multi-task training on our constructed large-scale dataset with 26.3 million samples, Nexus-Gen achieves state-of-the-art performance on the evaluation benchmarks spanning image understanding, generation and editing tasks. All models, datasets, and source codes are released in https://github.com/modelscope/Nexus-Gen to facilitate further advancements across the field.


\end{abstract}


\begin{figure*}[t] 
    \centering
    \includegraphics[width=1.\linewidth]{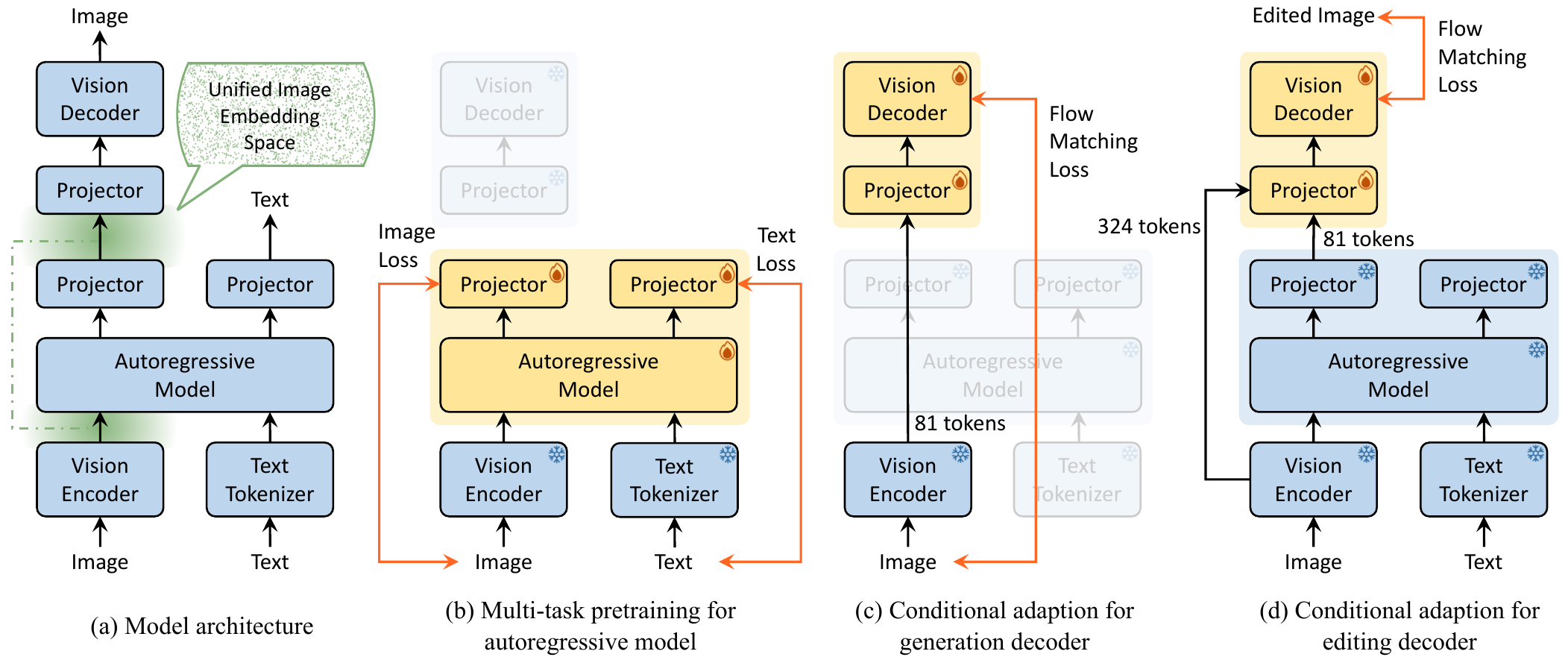}
    \caption{The architecture and the multi-stage training recipe for Nexus-Gen.}
    \label{fig:train_stage}
\end{figure*}

\section{Introduction}

Multimodal generative modeling has emerged as a pivotal frontier in AI research. Multimodal large language models (MLLMs) demonstrate notable competence in image understanding, while diffusion models lead in image generation. To further harness the potential of vision-language modeling, particularly the synergistic benefits across modalities, recent studies focused on MLLMs with unified modality architectures. The core advantage of unified modeling lies in the efficient data utilization and cross-modal representation learning, enabling the integration of multiple cross-modal capabilities. Such integration proves effective for complex tasks like image editing, visual reasoning, and reasoning-enhanced image generation, positioning unified MLLMs as a crucial pathway toward unified general intelligence.

Integrating image understanding and generation within a unified framework while enabling joint optimization remains a fundamental challenge for unified MLLMs. Prior works, including Chameleon \cite{chameleon}, Janus-Pro \cite{januspro}, and Emu3 \cite{wang2024emu3}, predominantly adopt autoregressive models coupled with variational autoencoders (VAEs) \cite{vae}. These frameworks employ autoregressive models to predict visual embeddings, which are fed into VQ-VAE \cite{vq} or VAE decoders for image generation. However, they underperform state-of-the-art diffusion models \cite{podell2023sdxl,esser2024sd3,flux} in image synthesis. This gap is attributed to the lack of pixel-level image modeling capabilities of autoregressive models. Another category of methods, such as SEED-X \cite{ge2024seedx} and MetaMerph \cite{metamorph}, predicts image embeddings autoregressively and employs diffusion models for image generation. A key limitation of these works is the unresolved error accumulation phenomenon during autoregressive token-by-token generation of continuous embeddings. Additionally, the extensibility of these frameworks to downstream tasks (e.g., image editing) remains insufficiently validated.

To leverage the rich pretrained knowledge of autoregressive LLMs and diffusion models, this paper proposes Nexus-Gen, a framework that unifies image understanding, generation, and editing tasks within a shared continuous image embedding space. As illustrated in Figure~\ref{fig:train_stage}(a), this embedding space serves as a pivotal interface between the autoregressive model and diffusion vision decoder. It also plays a vital role in preserving information integrity while maintaining embedding versatility across diverse tasks. For image understanding task, input images are encoded into the unified space to predict textual outputs. For image generation and editing tasks, the autoregressive model generates the target image embeddings within the space, which are subsequently decoded into images by vision decoders. Crucially, we uncover that autoregressive prediction of target image embeddings suffers from error severe accumulation. To address this, we propose the prefilled autoregression strategy, which prefills the input sequence with several learnable embeddings to align training and inference dynamics.

To perform joint optimization across multiple tasks and fully utilize the multimodal corpus, we construct a large-scale dataset of 26.3 million samples and propose a multi-stage training strategy for Nexus-Gen. The first stage conducts multi-task pretraining of the autoregressive model across image understanding, generation, and editing tasks. This process develops unified any-to-any modal prediction capabilities. The second stage adapts the generation vision decoder by fine-tuning it on a high-quality image generation dataset via replacing the textual conditioning with our unified image embeddings. The third stage adapts the editing decoder by fine-tuning it on our curated high-quality ImagePulse dataset to enable dual-stream image embedding inputs. Through multi-stage training, Nexus-Gen achieves superior performance on image understanding, generation and editing tasks. Specifically, it attains scores of 45.7 on the MMMU  understanding benchmark \cite{yue2023mmmu} and 0.81 on GenEval generation benchmark \cite{ghosh2023geneval}. Our contributions are as follows:

\begin{itemize}
    \item We propose Nexus-Gen, a unified model that leverages a unified image embedding space to bridge the capabilities of LLMs and diffusion models.
    \item We proposed a prefilling strategy that effectively avoids error accumulation during the prediction process of autoregressive models, thereby extending the generative capabilities of autoregressive models.
    \item We demonstrate the capabilities of Nexus-Gen through extensive experiments. Multiple benchmarks show that Nexus-Gen achieves state-of-the-art performance in image understanding, generation, and editing.
    \item We curate and release a dataset comprising 26.3 million samples for unified image understanding, generation, and editing tasks to promote research advances.
\end{itemize}

\section{Related Works}

Recent advances in unified architectures for image understanding and generation have stimulated research efforts, leading to two dominant paradigms: autoregressive model with VAE and autoregressive model with diffusion models.


\paragraph{Autoregressive Model with VAE} These methodologies exclusively employ lightweight VAEs \cite{vae} or VQ-VAEs \cite{vq} for image decoding, positioning the autoregressive model to modeling image information within the pixel space. Notably, Chameleon \cite{chameleon}, Show-O \cite{showo}, and Emu3 \cite{wang2024emu3} utilize a VQ-Tokenizer as the visual decoder, training the LLM on interleaved text-image data for unified modeling. Janus \cite{wu2025janus} and Janus-Pro \cite{januspro} further refine this architecture by decoupling understanding and generation tasks within different encoding spaces. They employ SigLIP \cite{siglip} for understanding-oriented encoding and VQ-Tokenizer for generative encoding, respectively. Crucially, all aforementioned methods rely on autoregressive prediction of subsequent visual tokens. Conversely, Transfusion \cite{transfusion}, and Janus Flow \cite{ma2025janusflow} adopt diffusion loss to optimize visual token generation.

\paragraph{Autoregressive Model with Diffusion} Leveraging pre-trained diffusion models as an additional component to synthesize image, these approaches typically yield superior image quality compared to autoregressive model with VAE techniques. The closed-source GPT-4o \cite{gpt4oimagegeneration} model exemplifies this architectural paradigm by adopting the workflow: token → [transformer] → [diffusion] → pixels. Representative open-source frameworks, including SEED-X \cite{ge2024seedx}, Emu2 \cite{emu2}, and MetaMerph \cite{metamorph}, adopt SDXL \cite{podell2023sdxl} as the vision decoder while utilizing regression loss to optimize the language LLM’s visual prediction capability. Regarding architectural variations under this paradigm, LM-Fusion \cite{shi2024llamafusion} and MetaQuery \cite{transfusion} both maintain the LLM in a frozen state. The former trains the vision decoder via shared attention mechanisms, while the latter employs learnable queries as an intermediary bridge between the LLM and the diffusion model. In contrast to the aforementioned methods, this work employs a unified embedding space to jointly model image understanding, generation and editing tasks. This design enables the LLM to capture cross-task correlations, facilitating subsequent research into interleaved tasks and reasoning-based multimodal understanding and generation. Additionally, we propose the prefilled autoregression strategy to optimize the generation of continuous image embeddings.


\section{Approach}

\subsection{Architecture}

The architecture of Nexus-Gen incorporates three core components, which are depicted in Figure~\ref{fig:train_stage}(a). The vision encoder and decoder are responsible for unified image embedding, while the autoregressive model facilitates unified multimodal context-aware reasoning.

\subsubsection{Unified Image Embedding Space}
Existing research \cite{gu2025breaking} on multimodal representation models demonstrates that unified embedding training across multiple downstream tasks facilitates a more comprehensive understanding of content information than single-task training. Thus, we propose a unified image embedding space to jointly model image-related tasks. For image understanding tasks, images are first encoded into image embeddings via a vision encoder, which are then further interpreted by the autoregressive model. For image generation task, the autoregressive model generates image embeddings based on textual descriptions, and the embeddings are subsequently decoded into images by the vision decoder. By integrating both understanding and generation capabilities, our framework enables image editing ability through modifications to the image embeddings. As unified models evolve towards reasoning-intensive and multi-turn conversational paradigms, our unified embedding space allows for the direct reuse of model-generated embeddings in subsequent reasoning or multi-turn conversations.

\subsubsection{Vision Encoder} 
We adopt the vision transformer of Qwen2.5-VL-7B-Instruct \cite{bai2025qwen2.5-vl} as our vision encoder and utilize its vision embedding space as our unified image embedding space. This space is implicitly aligned with textual representation space due to the multimodal abilities of the base model, which facilitates the establishment of robust text-image mapping with minimal training data. Leveraging the dynamic resolution capability of the vision encoder, we can modulate the number of image embedding tokens ($N_E$) by adjusting the input resolution ($H\times W$). This relationship is formally expressed as:
\begin{equation}
N_E = \left\lfloor \frac{H}{P} \right\rfloor \times \left\lfloor \frac{W}{P} \right\rfloor
\label{eq:token}
\end{equation}
where $P = 14$ denotes the size of each patch. Higher resolutions produce more token embeddings, with each token corresponding to a smaller spatial region and capturing more low-level features. Conversely, lower resolutions yield fewer tokens where each represents a larger spatial region, resulting in embeddings that encode more high-level features.

\subsubsection{Autoregressive Model}

The autoregressive model is utilized for unified multimodal reasoning, as illustrated in Figure~\ref{fig:train_stage}(a). Textual inputs are tokenized by the text tokenizer and projected into text embeddings, while visual inputs are encoded into image embeddings through the vision encoder. These text and image embeddings are jointly fed into the autoregressive model to predict the embeddings of output tokens. For image synthesis tasks, the output image embeddings predicted by the model are mapped to the unified image embedding space via a vision projector. For text generation, the output text embeddings are converted into logits by the text projector. In Nexus-Gen, the trainable parameters of the autoregressive model and text projector are initialized from Qwen2.5-VL-7B-Instruct \cite{bai2025qwen2.5-vl} to inherit the pretrained linguistic abilities. The vision projector is a randomly initialized linear layer for embedding alignment.

To prevent information loss, input images for understanding and editing tasks are encoded to embeddings at their native resolution without downsampling. For output images in generation and editing tasks, there exists a trade-off. Employing more image tokens helps capture finer image details. However, an excessive token count makes the task substantially more difficult for the autoregressive model, leading to degraded generation performance. Through experimental validation, we opted for a token count of 81, ensuring the reconstruction quality of the vision decoder without imposing excessive generation pressure on the autoregressive model.

\subsubsection{Vision Decoder}


To achieve high-fidelity image decoding from model-generated embeddings, we adopt the diffusion transformer of FLUX.1-Dev \cite{flux} as our vision decoder by replacing its native T5 text embeddings \cite{t5} with our designed conditioning mechanisms. Given the divergent objectives of image generation and editing tasks, we implement specialized conditioning schemes and architectural configurations for the respective decoders.

For the image generation task, which emphasizes semantic consistency with textual descriptions, the decoder is designed to reconstruct images that are semantically consistent with the 81-token image embeddings produced by the autoregressive model, as presented in Figure \ref{fig:train_stage}(c). To condition the diffusion transformer on these embeddings, a two-layer MLP projector is utilized for embedding alignment.

For image editing tasks requiring faithful execution of editing instructions while preserving details in unaltered regions, we propose an editing decoder with dual conditioning mechanisms, as detailed in Figures~\ref{fig:train_stage}(d). The first condition incorporates 81-token embeddings from the autoregressive model, conveying semantic information of the target image. The second condition integrates 324-token embeddings derived from direct encoding of the input image, preserving fine-grained details of the original content. To effectively model the hierarchical relationships between these two conditions, we introduce a joint attention layer as the embedding projector for the diffusion transformer.

\begin{figure}[t] 
    \centering
    \includegraphics[width=1.\linewidth]{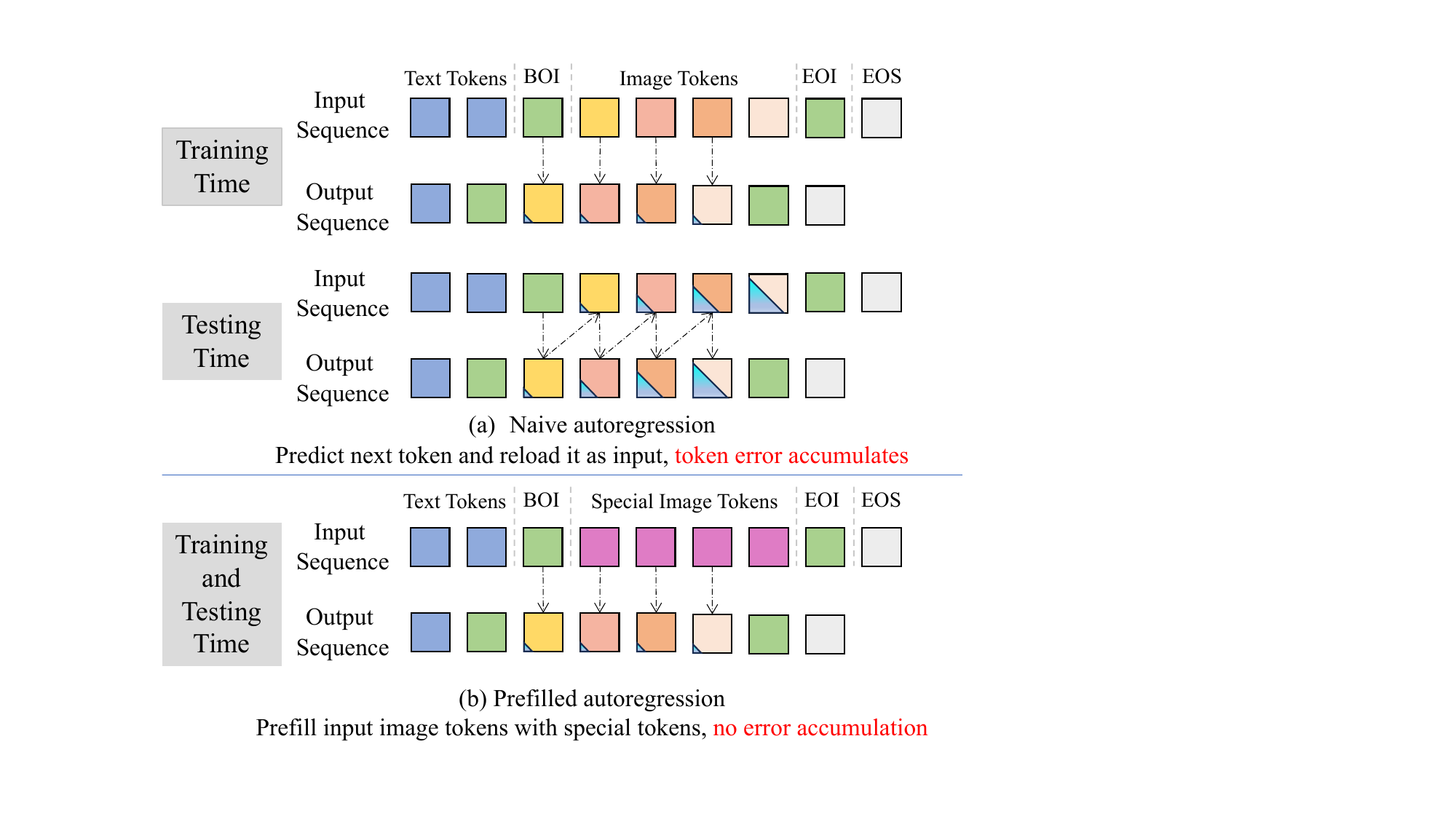}
    \caption{(a) The naive autoregressive approach exhibits inconsistent behaviors between training and test phases, leading to error accumulation during inference. (b) We propose a novel strategy that prefills special image tokens during training and testing, which unifies the computational behaviors across both phases and eliminates error accumulation.}
    \label{fig:ar_prefill}
\end{figure}

\subsection{Prefilled Autoregression}

When optimizing the autoregressive model, we observe significant error accumulation in the continuous embedding space, attributed to the discrepancy between the training and inference behaviors of autoregressive models. As depicted in Figure~\ref{fig:ar_prefill}(a), the model leverages ground-truth preceding tokens for each prediction during teacher-forcing training, whereas at inference time, it relies on autoregressively generated tokens. When processing continuous image embeddings, the model directly predicts biased embeddings and feeds them back as input. This recursive injection of prediction errors propagates and amplifies across subsequent tokens, resulting in suboptimal performance.

Prior research \cite{mar} demonstrates that image token prediction is permutation-invariant. This suggests that each image embedding can be derived solely from the text caption and positional encoding without depending on preceding embeddings. Thus, we propose the prefilled autoregression strategy to mitigate error accumulation, illustrated in Figure~\ref{fig:ar_prefill}(b). During training, we initialize all image tokens with $N_E$ learnable embeddings with positional encoding (where $N_E$ is the preset token quantity). During inference, upon predicting the BOI (beginning of image) token, we prefill the input sequence with the learned embeddings. This enforces alignment between training and inference by preventing error-affected predictions from being recycled into the input, thereby eliminating error accumulation.


\begin{table*}[thp]
\centering
\begin{tabular}{lccccccc}
\toprule
Model          & MME-P $\uparrow$       & MME-C $\uparrow$ & SEED $\uparrow$ & MMMU $\uparrow$ & TextVQA $\uparrow$ & VQAv2 $\uparrow$ & RWQA $\uparrow$       \\ \midrule
Qwen2.5-VL-Instruct 7B $^\dagger$ & 1689.0          & 640.3          & 77.4          & 50.6          & 77.9          & 82.3          & 66.0          \\ \midrule
EMU2 Chat 34B  & -            & -     & 62.8 & 34.1 & \underline{66.6}    & -     & -          \\
Seed-X 17B     & 1457.0       & -     & 66.5 & 35.6 & -       & 63.4  & -          \\
Chameleon 7B    & 202.7        & -     & 27.2 & 22.4 & -          & -     & 39.0       \\
Chameleon 34B  & 604.5        & -     & -    & 38.8 & -       & 69.6  & 39.2       \\
EMU3 8B        & 1243.8       & 266.1 & 68.2 & 31.6 & 64.7    & 75.1  & 57.4       \\
MetaMorph 8B   & -            & -     & 71.4 & 41.8 & 60.5    & -     & 58.3       \\
Show-O 1.3B    & 1097.2       & -     & -    & 27.4 & -       & 69.4  & -          \\
VILA-U 7B       & 1336.2       & -     & 56.3 & 32.2 & 48.3       & 75.3  & 46.6       \\
Janus 1.5B     & 1338.0       & -     & 63.7 & 30.5 & -       & -     & -          \\
Janus-Pro 1.5B & 1444.0       & -     & 68.3 & 36.3 & -       & -     & -          \\
Janus-Pro 7B   & \underline{1567.1} & -     & 72.1 & 41.0 & -       & -     & 60.0 \\
LMFusion 16B   & 1603.7       & 367.8 & 72.1 & 41.7 & -       & -     & \underline{60.0} \\
TokenFlow-L 13B & 1365.4       & 257.5 & 62.6 & 34.4 & 54.1       & 73.9  & 49.2       \\
TokenFlow-XL 14B       & 1551.1          & \underline{371.1}    & \underline{72.6}    & \underline{43.2}    & 62.3          & \underline{77.6}    & 56.6          \\ \midrule
Nexus-Gen 7B (Ours)          & \textbf{1602.3} & \textbf{637.5} & \textbf{77.1} & \textbf{45.7} & \textbf{75.5}    & \textbf{79.3} & \textbf{63.7} \\
\bottomrule
\end{tabular}
\caption{Evaluation on image understanding benchmarks. We highlight the best results in bold and underline the second-best result. $\dagger$ We evaluate our MLLM base model Qwen2.5-VL-Instruct 7B as a reference baseline.}
\label{tab:image understanding}
\end{table*}

\subsection{Dataset Curation}
\label{sec:dataset curation}
To enable Nexus-Gen with unified visual capabilities, we construct a dataset with 26.3 million samples covering image understanding, generation, and editing tasks. The majority of our dataset derives from publicly accessible open-source repositories \cite{zhang2025eligen, cambrian, zhao2024ultraedit}, which is relabeled to improve annotation quality. However, given the image quality limitations (e.g., aesthetic, artifacts) in existing editing datasets, we additionally construct a high-quality image editing dataset, ImagePulse. We will release all training data after data security and legality checks. The complete dataset construction pipeline is provided in the Appendix.

\subsection{Training Objectives}
The training of Nexus-Gen encompasses both its autoregressive model and vision decoder components. The autoregressive model undergoes unified multi-task training, generating outputs comprising both text and image embeddings. The loss function for the text and image embeddings is formulated as follows, where $\lambda_1=3$, $\lambda_2=1.5$, $\lambda_3=1.5$ are hyperparameters that control the loss weights.
\begin{align}
    L&= L_{\text {Text}} + L_{\text {Image}} \\
    &=\lambda_1  \cdot L_{\text {CE}} + (\lambda_2 \cdot L_{\text {MSE}} + \lambda_3 \cdot L_{\text {COS}})
\end{align}

For the text tokens, we employ the standard cross-entropy loss for classification, which is defined in Eq.~\ref{eq:ce}. Here, $N_T$ denotes text tokens numbers, $|V|$ represents the vocabulary size, $y_t^{(c)}$ is the ground-truth one-hot encoded token, and $\hat{y}_t^{(c)}$ indicates the predicted probability distribution.
\begin{equation}
L_{\text {CE}} = -\frac{1}{N_T} \sum_{t=1}^{N_T} \sum_{c=1}^{|V|} y_t^{(c)} \log(\hat{y}_t^{(c)}) \label{eq:ce}
\end{equation}

For the image embeddings, we utilize a composite loss function combining mean squared error and cosine similarity loss \cite{clip}. This combination ensures the preservation of detail fidelity while simultaneously enforcing semantic coherence. The loss functions are defined in Eq.~\ref{eq:mse} and~\ref{eq:cos}, where $N_E$ is image embeddings numbers, $D$ signifies the embedding dimensionality, $\hat{\mathbf{e}}_i$ and $\mathbf{e}_i$  denote the predicted and ground-truth embeddings, respectively.
\begin{equation}
    L_{\text {MSE}} = \frac{1}{N_E \cdot D} \sum_{i=1}^{N_E} \Vert \mathbf{e}_i - \hat{\mathbf{e}}_i \Vert_2^2 \label{eq:mse}
\end{equation}
\begin{equation}
    L_{\text {COS}}= - \frac{1}{N_E} \sum_{i=1}^{N_E} \frac{\mathbf{e}_i \cdot \hat{\mathbf{e}}_i}{\Vert \mathbf{e}_i \Vert_2 \cdot \Vert \hat{\mathbf{e}}_i \Vert_2} \label{eq:cos}
\end{equation}

The vision decoders for generation and editing tasks are trained separately. They both adopt the MSE loss function of flow matching. Given the diffusion transformer $V_\theta$, target image $X_1$ in latent space, noise $X_0\sim N(0,1)$, conditions $C$ and timestep $t\sim \mu(0, 1)$, the loss function is defined as:

\begin{equation}
L_{\text {Flow }}=\mathbb{E}\left[\left\|V_\theta\left(X_t, C, t\right)-(X_1-X_0)\right\|^2\right]
\end{equation}


\subsection{Training Strategy}
We adopt a multi-stage training strategy for Nexus-Gen. The first stage consists of the unified multi-task pretraining and the aesthetic fine-tuning for autoregressive model, as is shown in Figure~\ref{fig:train_stage}(b). The second and third stages conduct conditional adaptation for the generation decoder and editing decoder, which is illustrated in Figure~\ref{fig:train_stage}(c) and (d). All training hyperparameters are listed in the Appendix.

\subsubsection{Multi-Task Pretraining for Autoregressive Model}
The first stage executes unified pretraining and aesthetic fine-tuning of the autoregressive model. Pretraining utilizes the complete dataset of 26.3 million samples, primarily preserving the autoregressive model's inherent text prediction capacity while establishing a visual embedding prediction functionality. Subsequent aesthetic fine-tuning process employs 4.3 million high-quality samples to optimize Nexus-Gen's visual output quality. This high-quality dataset comprises: (1) premium image generation samples \cite{zhang2025eligen,flux_aes,flux_T2I,chen2025blip3}, (2) our novel ImagePulse dataset augmented with 0.5 million randomly selected instances from other editing corpora \cite{zhao2024ultraedit,hqedit}, and (3) 1 million image understanding data \cite{cambrian}.


\subsubsection{Conditional Adaption for Generation Decoder}
The second stage fine-tunes the generation decoder to harmonize its conditional inputs with the unified image embedding space via an image reconstruction objective. To preserve visual fidelity, this module is exclusively trained on two million high-quality image generation samples \cite{zhang2025eligen,flux_aes,flux_T2I,chen2025blip3}.

\subsubsection{Conditional Adaption for Editing Decoder}
The third stage fine-tunes the editing decoder using our ImagePulse dataset, as illustrated in Figure~\ref{fig:train_stage}(d). This decoder processes dual heterogeneous inputs: 324-token fine-grained embeddings extracted from the input image and 81-token semantic embeddings generated by the autoregressive model.

\section{Experiments}
\subsection{Main Results}
In this section, we conduct a quantitative evaluation of Nexus-Gen across multiple benchmarks for image understanding, generation, and editing tasks.
\subsubsection{Image Understanding}
For image understanding tasks, we evaluate the model performance on several multimodal understanding benchmarks: MME \cite{fu2024mmecomprehensiveevaluationbenchmark}, SEEDBench \cite{li2023seed}, MMMU \cite{yue2023mmmu}, TextVQA \cite{textvqa}, VQAv2 \cite{vqa}, and RealWorldQA \cite{rwqa}. For baseline comparison, we incorporate the following unified models: EMU2 \cite{emu2}, Seed-X \cite{ge2024seedx}, Chameleon \cite{chameleon}, Emu3 \cite{wang2024emu3}, Metamorph \cite{metamorph}, Show-O \cite{metamorph}, VILA-U \cite{wu2024vila}, Janus \cite{wu2025janus}, Janus-Pro \cite{januspro}, LMFusion \cite{shi2024llamafusion}, and TokenFlow \cite{qu2025tokenflow}. As is shown in Table~\ref{tab:image understanding}, the 7B-parameter Nexus-Gen achieves state-of-the-art performance across all evaluated benchmarks. It outperforms other unified models by significant margins, with particular advantages on MME-Cognition and TextVQA. We further evaluated our MLLM baseline Qwen2.5-VL-Instruct 7B. Empirical results demonstrate that Nexus-Gen endows the base autoregressive model with image generation and editing capabilities without incurring significant loss in image understanding performance, while introducing no additional parameters.

\begin{table*}[t]
\centering
\scalebox{0.98}{
\begin{tabular}{lccccccc}
\toprule
Method         & Single Object $\uparrow$ & Two Object $\uparrow$ & Counting $\uparrow$ & Colors $\uparrow$ & Position $\uparrow$ & Color Attribute $\uparrow$ & Overall $\uparrow$      \\ \midrule
Emu3-Gen       & 0.99        & 0.81     & 0.42     & 0.80    & 0.49     & 0.45         & 0.66          \\
SEED-X         & 0.97        & 0.58     & 0.26     & 0.80    & 0.19     & 0.14         & 0.49          \\
Transfusion    & -           & -        & -        & -      & -        & -            & 0.63          \\
Show-O         & 0.95        & 0.52     & 0.49     & 0.82   & 0.11     & 0.28         & 0.53          \\
MetaQuery-XL   & -           & -        & -        & -      & -        & -            & 0.80          \\
TokenFlow-XL   & 0.95        & 0.60      & 0.41     & 0.81   & 0.16     & 0.24         & 0.55          \\
Janus          & 0.97        & 0.68     & 0.3      & 0.84   & 0.46     & 0.42         & 0.61          \\
Janus-Pro 1.5B & 0.98        & 0.82     & 0.51     & \underline{0.89}   & 0.65     & 0.56         & 0.73          \\
Janus-Pro 7B   & \textbf{0.99}        & \underline{0.89}     & \underline{0.59}     & \textbf{0.90}   & \underline{0.79}     & \textbf{0.66}         & \underline{0.80}          \\ \midrule
Nexus-Gen (Ours)      & \underline{0.99}        & 0.86     & 0.53     & 0.85   & 0.78     & 0.59         & 0.77          \\
Nexus-Gen* (Ours)     & 0.97        & \textbf{0.93}     & \textbf{0.64}     & 0.88   & \textbf{0.83}     & \underline{0.62}         & \textbf{0.81} \\ \bottomrule
\end{tabular}
}
\caption{Evaluation of image generation ability on GenEval benchmark. We highlight the best results in bold, and underline the second best result. Nexus-Gen underwent joint optimization across all three tasks. We subsequently perform instruction tuning on the Blip3o-60k dataset targeting image generation task, resulting in the optimized Nexus-Gen* variant.}
\label{tab:image generation}
\end{table*}

\begin{table}[thp]
\centering
\scalebox{0.9}{
\begin{tabular}{lccccc}
\toprule
Method & CLIP-T ↑ & L1 ↓ & CLIP-O ↑ & DINO-O ↑ \\ \midrule
InstructPix2Pix & 0.299 & 0.171 & 0.832 & 0.706 \\
MagicBrush & 0.309 & 0.146 & 0.863 & 0.750 \\
AnyEdit & 0.305 & \underline{0.141} & 0.863 & 0.756 \\
UltraEdit & 0.306 & 0.157 & 0.841 & 0.737 \\
OmniGen & 0.317 & 0.154 & 0.874 & 0.764 \\
Step1X-Edit & \underline{0.317} & 0.142 & \underline{0.879} & \underline{0.779} \\ \midrule
Nexus-Gen & \textbf{0.324} & \textbf{0.134} & \textbf{0.909} & \textbf{0.834} \\ \bottomrule
\end{tabular}
}
\caption{Evaluation of image editing ability on ImagePulse benchmark. We highlight the best results in bold, and underline the second best result.}
\label{tab:image editing}
\end{table}

\subsubsection{Image Generation}
For text-to-image generation tasks, we adopt GenEval \cite{ghosh2023geneval} as the evaluation benchmark, with assessment metrics covering object fidelity, quantity accuracy, color correctness, attribute matching, and spatial relationships. We compare Nexus-Gen and its instruction-tuned variant Nexus-Gen* (fine-tuned on Blip3o-60k dataset \cite{chen2025blip3}) against unified models including EMU3, SEED-X, Transfusion \cite{transfusion}, Show-O \cite{showo}, MetaQuery \cite{pan2025transfer}, TokenFlow \cite{qu2025tokenflow}, Janus \cite{wu2025janus}, and Janus-Pro \cite{januspro}. Results in Table~\ref{tab:image generation} reveal that the multi-task jointly-trained Nexus-Gen achieves an overall score of 0.77. After generative-task-specific instruction tuning, Nexus-Gen* attains state-of-the-art performance with a score of 0.81. This enhanced model demonstrates significant advantages in Two Object, Counting and Position metrics compared to baseline methods.

\subsubsection{Image Editing}
For image editing evaluation, we utilize the test set with 1,000 randomly sampled cases from ImagePulse dataset (non-overlapping with training data). On this benchmark, we compare Nexus-Gen against state-of-the-art editing models: InstructPix2Pix \cite{brooks2023instructpix2pix}, MagicBrush \cite{zhang2023magicbrush}, AnyEdit \cite{yu2025anyedit}, UltraEdit \cite{zhao2024ultraedit}, OmniGen \cite{xiao2025omnigen}, and Step1X-Edit \cite{liu2025step1x}. We employ three complementary metric categories: (1) CLIP-T measures the CLIP image-text similarity between edited image and target caption. (2) L1 measures the pixel-level absolute difference between the edited image and groundtruth image. (3) CLIP-O and DINO-O measure the cosine similarity between the edited image and groundtruth image using their CLIP \cite{clip} and DINO \cite{dino} embeddings. As evidenced in Table~\ref{tab:image editing}, Nexus-Gen demonstrates consistent superiority across all metrics. This performance advantage indicates Nexus-Gen's enhanced capability to faithfully execute editing instructions and generate outputs with significantly improved alignment to both target captions and ground truth images.

\begin{figure}[thp] 
    \centering
    \includegraphics[width=1.\linewidth]{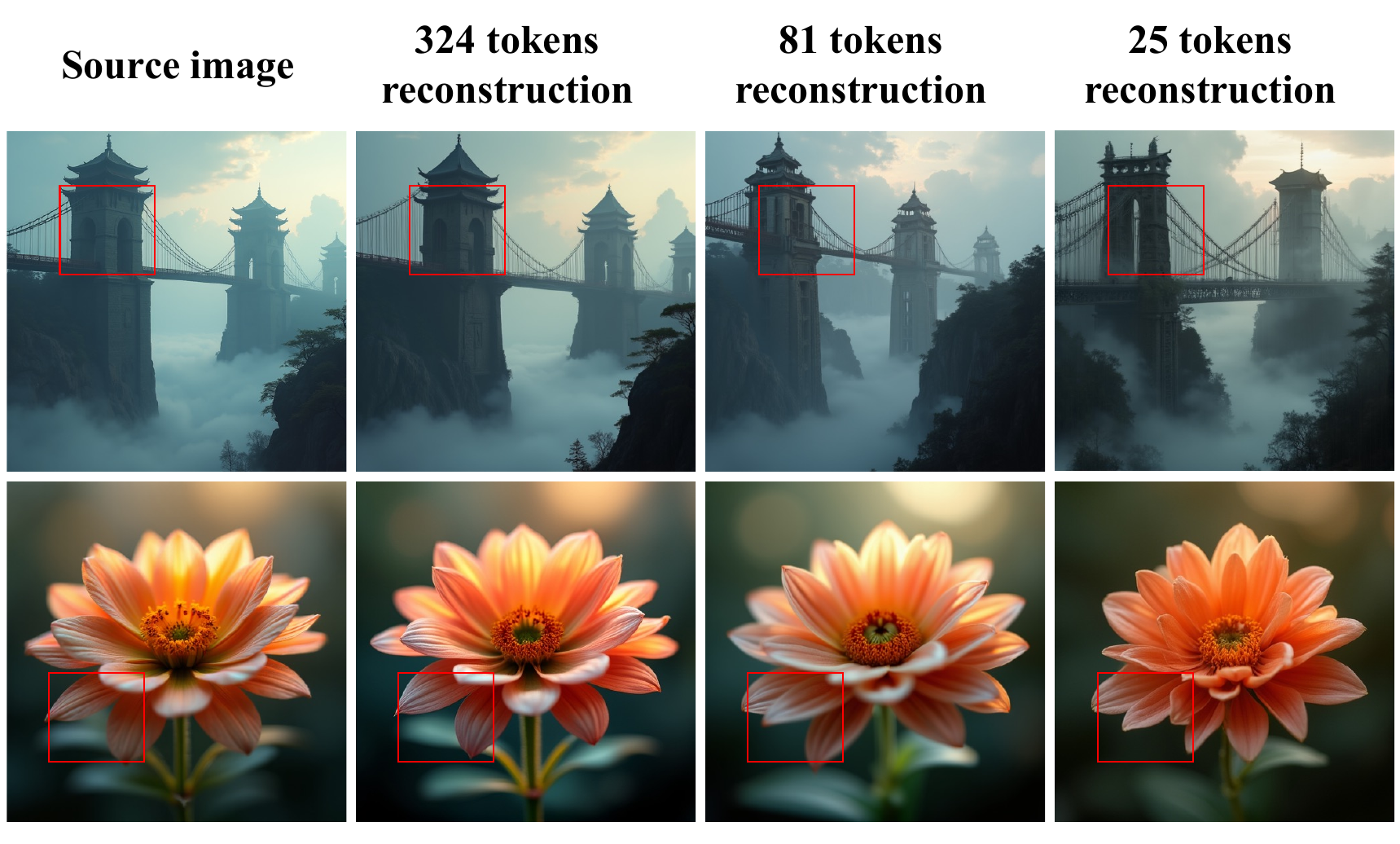}
    \caption{Image reconstruction results of our generation decoder using 81 and 324 image token embedding.}
    \label{fig:reconstruction}
\end{figure}

\begin{figure}[thp] 
    \centering
    \includegraphics[width=1.\linewidth]{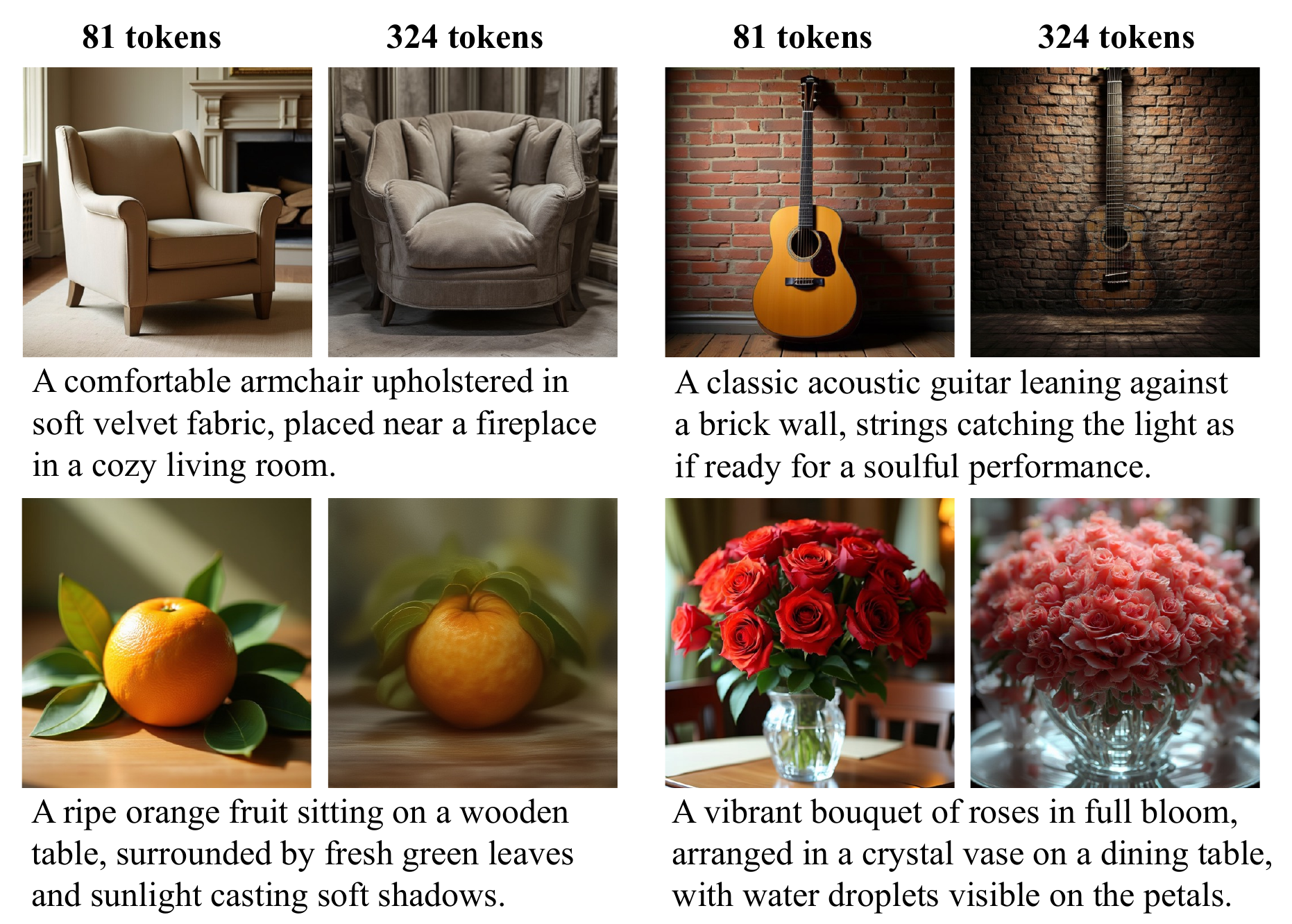}
    \caption{Image generation results from Nexus-Gen trained with 324 and 81 image token embeddings.}
    \label{fig:llm_81_324}
\end{figure}

\begin{figure}[thp] 
    \centering
    \includegraphics[width=1.\linewidth]{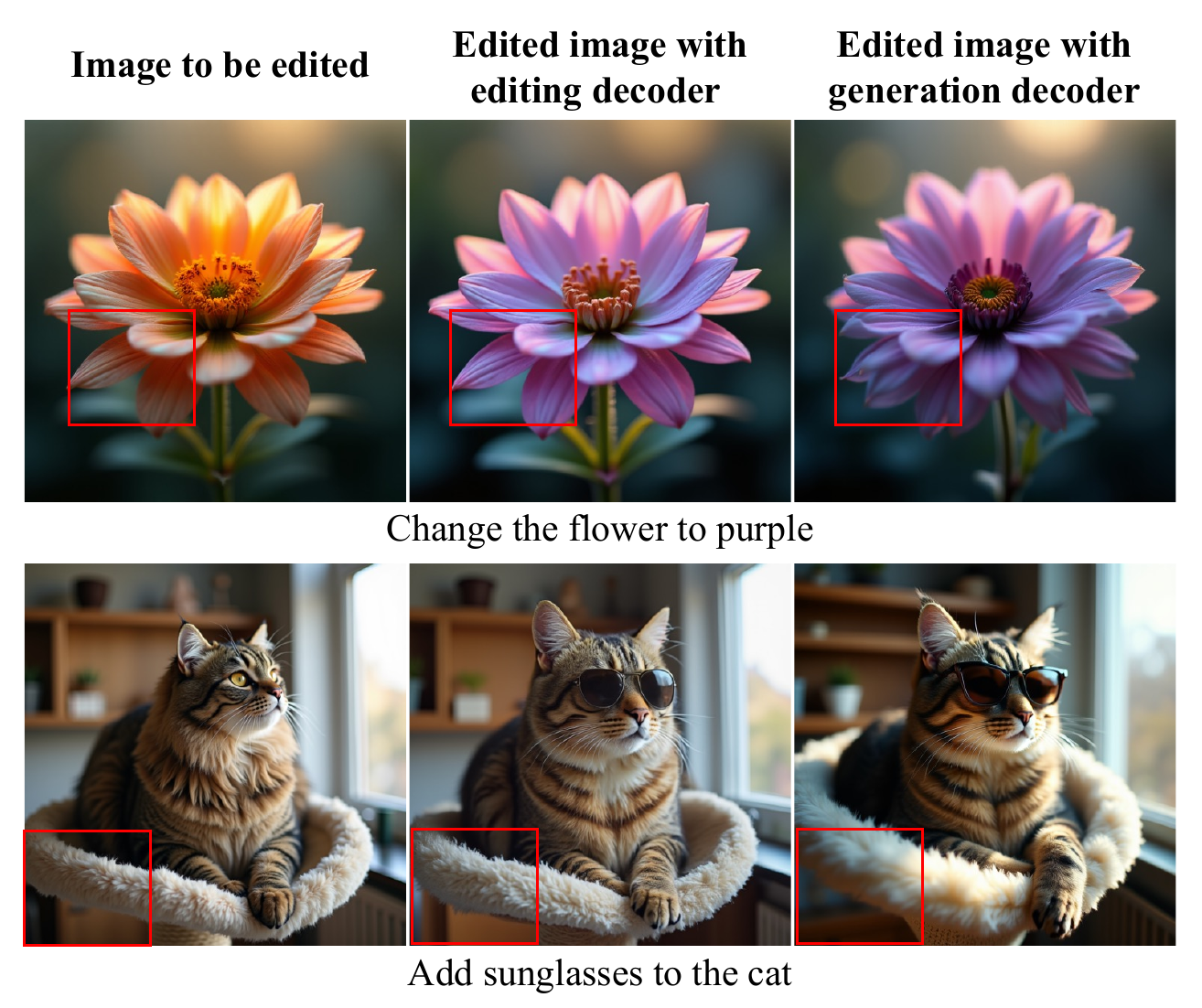}
    \caption{Image editing results of Nexus-Gen with editing and generation decoder.}
    \label{fig:editing_results}
\end{figure}

\subsection{Ablation Studies}
\subsubsection{Trade-offs in Token Quantity}
The token quantity for image embeddings exhibits a positive correlation with resolution. Higher resolution produces more tokens, allowing embeddings to retain finer visual details. We consider three standard resolutions: $128\times128$, $256\times256$ and $512\times512$, corresponding to 25, 81 and 324 tokens, respectively. We train the generation decoder at these token quantities for image reconstruction, with qualitative results shown in Figure~\ref{fig:reconstruction}. The reconstructions with 81 and 324 tokens successfully preserve global layouts and high-level semantics of source images. Notably, the 324-token approach demonstrates significantly superior detail consistency. In contrast, the 25-token reconstruction exhibits structural distortions and semantic loss relative to the source image.

We further train the autoregressive model at token counts of 81 and 324 and validate image generation quality using above vision decoders. Experimental results in Figure~\ref{fig:llm_81_324} demonstrate that the model trained with 81 tokens effectively generates images aligned with textual semantics. However, the 324-token model exhibits severe semantic repetition in generated images, which exhibited compromised quality. This indicates that autoregressive models struggle to accurately predict such extensive image tokens. Consequently, we select token quantity of 81 as the optimal count for both the autoregressive model and generation decoder.

\begin{figure}[tp] 
    \centering
    \includegraphics[width=1.\linewidth]{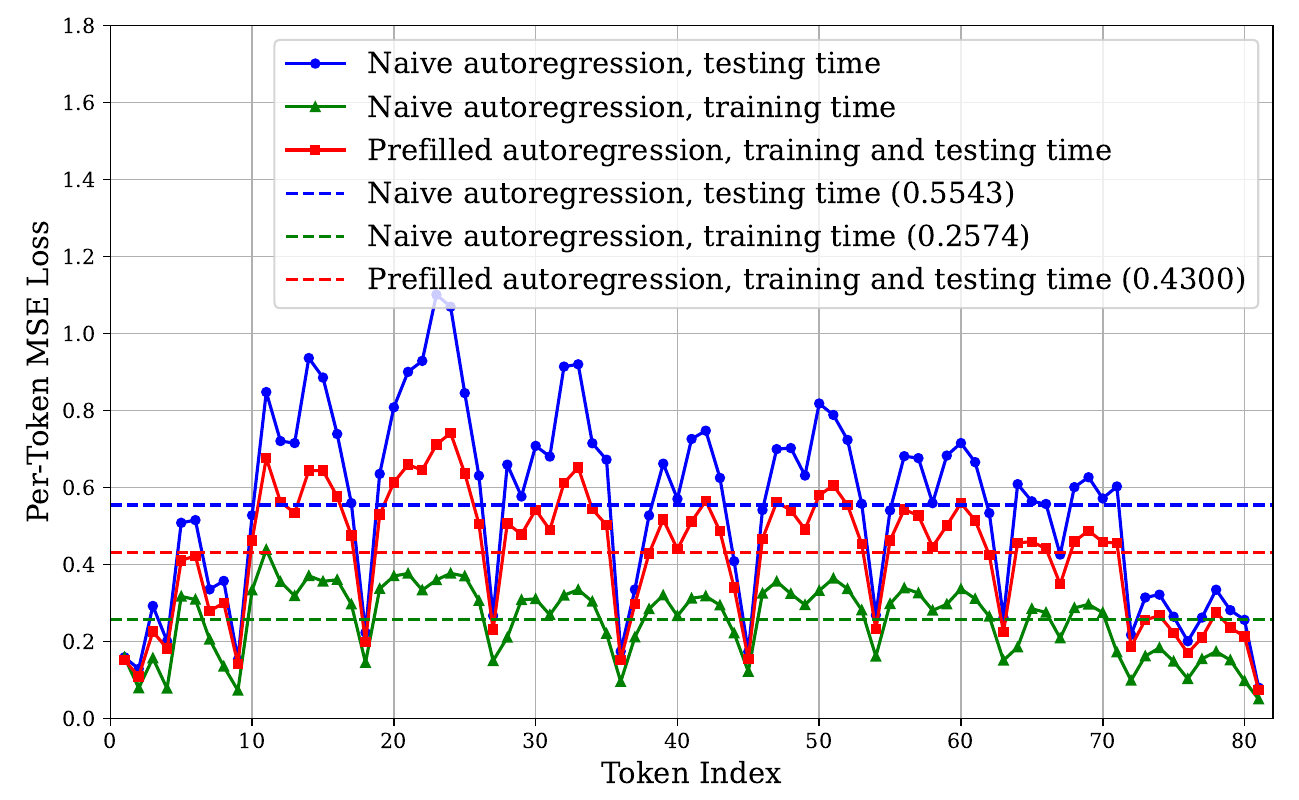}
    \caption{MSE loss comparison between the naive and prefilled autoregression strategy.}
    \label{fig:loss_comparison}
\end{figure}

\subsubsection{The Necessity of Editing Decoder}
Although the generation decoder can be directly applied to image editing tasks, we propose the editing decoder (Figure~\ref{fig:train_stage}d) to enhance the detail preservation capability in non-edited regions. Figure~\ref{fig:editing_results} compares the editing performance of both decoders. Given identical inputs, both solutions successfully execute edit instructions. However, the generation decoder fails to maintain details in non-edited areas due to its 81-token reconstruction constraints. In contrast, the editing decoder synergizes the superior 324-token reconstruction capability with the efficient 81-token embedding prediction, achieving demonstrably higher editing fidelity.

\subsubsection{The Impact of Prefilled Autoregression}
To mitigate the error accumulation issue in naive autoregression paradigm, we propose the prefilled autoregression strategy. Figure~\ref{fig:loss_comparison} compares the MSE losses of image tokens predicted by both approaches. During training, naive autoregression achieves the lowest loss of 0.2574 due to access to preceding ground-truth embedding. During inference however, the autoregressive nature causes progressive error accumulation, elevating the average loss to 0.5543. In contrast, our prefilled autoregression strategy maintains consistent training-inference behavior, achieving a significantly lower inference loss of 0.43.

\section{Conclusion and Future Works}
In this work, we present Nexus-Gen, a unified model for image understanding, generation, and editing tasks. The core innovation of Nexus-Gen lies in bridging the language reasoning capabilities of LLMs with the image synthesis power of diffusion models through a unified continuous image embedding space. Furthermore, we identify the error accumulation phenomenon during the autoregressive prediction of continuous embeddings and propose prefilled autoregression strategy to mitigate it. To perform joint optimization across multiple tasks, we curate a large-scale dataset of 26.3 million samples and train Nexus-Gen using a multi-stage strategy, which includes the multi-task pretraining of the autoregressive model and conditional adaptations of the generation and editing decoders. Extensive experiments validate Nexus-Gen’s state-of-the-art performance across all tasks. While Nexus-Gen successfully unifies the three image tasks, certain limitations warrant attention. The model exhibits compromised robustness to prompt variations during image generation, and more significantly, its capacity for sophisticated visual reasoning remains unexplored. To address these constraints, we will focus on developing Nexus-Gen’s advanced applications in complex tasks such as in-context learning and step-by-step vision-language reasoning.

\bibliography{aaai2026}

\begin{thebibliography}{57}
\providecommand{\natexlab}[1]{#1}

\bibitem[{Bai et~al.(2025)Bai, Chen, Liu, Wang, Ge, Song, Dang, Wang, Wang, Tang et~al.}]{bai2025qwen2.5-vl}
Bai, S.; Chen, K.; Liu, X.; Wang, J.; Ge, W.; Song, S.; Dang, K.; Wang, P.; Wang, S.; Tang, J.; et~al. 2025.
\newblock Qwen2. 5-vl technical report.
\newblock \emph{arXiv preprint arXiv:2502.13923}.

\bibitem[{Brooks, Holynski, and Efros(2023)}]{brooks2023instructpix2pix}
Brooks, T.; Holynski, A.; and Efros, A.~A. 2023.
\newblock Instructpix2pix: Learning to follow image editing instructions.
\newblock In \emph{Proceedings of the IEEE/CVF conference on computer vision and pattern recognition}, 18392--18402.

\bibitem[{Caron et~al.(2021)Caron, Touvron, Misra, J\'egou, Mairal, Bojanowski, and Joulin}]{dino}
Caron, M.; Touvron, H.; Misra, I.; J\'egou, H.; Mairal, J.; Bojanowski, P.; and Joulin, A. 2021.
\newblock Emerging Properties in Self-Supervised Vision Transformers.
\newblock In \emph{Proceedings of the International Conference on Computer Vision (ICCV)}.

\bibitem[{Chen et~al.(2025{\natexlab{a}})Chen, Xu, Pan, Hu, Qin, Goldstein, Huang, Zhou, Xie, Savarese et~al.}]{chen2025blip3}
Chen, J.; Xu, Z.; Pan, X.; Hu, Y.; Qin, C.; Goldstein, T.; Huang, L.; Zhou, T.; Xie, S.; Savarese, S.; et~al. 2025{\natexlab{a}}.
\newblock Blip3-o: A family of fully open unified multimodal models-architecture, training and dataset.
\newblock \emph{arXiv preprint arXiv:2505.09568}.

\bibitem[{Chen et~al.(2025{\natexlab{b}})Chen, Wu, Liu, Pan, Liu, Xie, Yu, and Ruan}]{januspro}
Chen, X.; Wu, Z.; Liu, X.; Pan, Z.; Liu, W.; Xie, Z.; Yu, X.; and Ruan, C. 2025{\natexlab{b}}.
\newblock Janus-Pro: Unified Multimodal Understanding and Generation with Data and Model Scaling.
\newblock \emph{arXiv preprint arXiv:2501.17811}.

\bibitem[{Creative(2024)}]{FLUX-Controlnet}
Creative, A. 2024.
\newblock FLUX-Controlnet-Inpainting.
\newblock \url{https://github.com/alimama-creative/FLUX-Controlnet-Inpainting.git}.

\bibitem[{Esser et~al.(2024)Esser, Kulal, Blattmann, Entezari, M{\"u}ller, Saini, Levi, Lorenz, Sauer, Boesel et~al.}]{esser2024sd3}
Esser, P.; Kulal, S.; Blattmann, A.; Entezari, R.; M{\"u}ller, J.; Saini, H.; Levi, Y.; Lorenz, D.; Sauer, A.; Boesel, F.; et~al. 2024.
\newblock Scaling rectified flow transformers for high-resolution image synthesis.
\newblock \emph{arXiv preprint arXiv:2403.03206}.

\bibitem[{Fu et~al.(2024)Fu, Chen, Shen, Qin, Zhang, Lin, Yang, Zheng, Li, Sun, Wu, and Ji}]{fu2024mmecomprehensiveevaluationbenchmark}
Fu, C.; Chen, P.; Shen, Y.; Qin, Y.; Zhang, M.; Lin, X.; Yang, J.; Zheng, X.; Li, K.; Sun, X.; Wu, Y.; and Ji, R. 2024.
\newblock MME: A Comprehensive Evaluation Benchmark for Multimodal Large Language Models.
\newblock arXiv:2306.13394.

\bibitem[{Ge et~al.(2024)Ge, Zhao, Zhu, Ge, Yi, Song, Li, Ding, and Shan}]{ge2024seedx}
Ge, Y.; Zhao, S.; Zhu, J.; Ge, Y.; Yi, K.; Song, L.; Li, C.; Ding, X.; and Shan, Y. 2024.
\newblock Seed-x: Multimodal models with unified multi-granularity comprehension and generation.
\newblock \emph{arXiv preprint arXiv:2404.14396}.

\bibitem[{Ghosh, Hajishirzi, and Schmidt(2023)}]{ghosh2023geneval}
Ghosh, D.; Hajishirzi, H.; and Schmidt, L. 2023.
\newblock Geneval: An object-focused framework for evaluating text-to-image alignment.
\newblock \emph{Advances in Neural Information Processing Systems}, 36: 52132--52152.

\bibitem[{gogoduan(2025)}]{flux_aes}
gogoduan. 2025.
\newblock flux-laion-aes.

\bibitem[{Goyal et~al.(2017)Goyal, Khot, Summers-Stay, Batra, and Parikh}]{vqa}
Goyal, Y.; Khot, T.; Summers-Stay, D.; Batra, D.; and Parikh, D. 2017.
\newblock Making the v in vqa matter: Elevating the role of image understanding in visual question answering.
\newblock In \emph{Proceedings of the IEEE conference on computer vision and pattern recognition}, 6904--6913.

\bibitem[{Gu et~al.(2025)Gu, Yang, Feng, Wang, Zhang, Long, Chen, Cai, and Deng}]{gu2025breaking}
Gu, T.; Yang, K.; Feng, Z.; Wang, X.; Zhang, Y.; Long, D.; Chen, Y.; Cai, W.; and Deng, J. 2025.
\newblock Breaking the Modality Barrier: Universal Embedding Learning with Multimodal LLMs.
\newblock \emph{arXiv preprint arXiv:2504.17432}.

\bibitem[{Han et~al.(2024)Han, Mao, Jiang, Pan, and Zhang}]{han2024stylebooth}
Han, Z.; Mao, C.; Jiang, Z.; Pan, Y.; and Zhang, J. 2024.
\newblock StyleBooth: Image Style Editing with Multimodal Instruction.
\newblock \emph{arXiv preprint arXiv:2404.12154}.

\bibitem[{Hui et~al.(2024)Hui, Yang, Zhao, Shi, Wang, Wang, Zhou, and Xie}]{hqedit}
Hui, M.; Yang, S.; Zhao, B.; Shi, Y.; Wang, H.; Wang, P.; Zhou, Y.; and Xie, C. 2024.
\newblock Hq-edit: A high-quality dataset for instruction-based image editing.
\newblock \emph{arXiv preprint arXiv:2404.09990}.

\bibitem[{jackyhate(2024)}]{flux_T2I}
jackyhate. 2024.
\newblock text-to-image-2M.

\bibitem[{Jiang et~al.(2025)Jiang, Yan, Jia, Liu, Kang, and Lu}]{jiang2025infiniteyou}
Jiang, L.; Yan, Q.; Jia, Y.; Liu, Z.; Kang, H.; and Lu, X. 2025.
\newblock {InfiniteYou}: Flexible Photo Recrafting While Preserving Your Identity.
\newblock In \emph{ICCV}.

\bibitem[{Kingma, Welling et~al.(2013)}]{vae}
Kingma, D.~P.; Welling, M.; et~al. 2013.
\newblock Auto-encoding variational bayes.

\bibitem[{Labs(2024)}]{flux}
Labs, B.~F. 2024.
\newblock FLUX.
\newblock \url{https://blackforestlabs.ai/announcing-black-forest-labs}.

\bibitem[{LAION(2024)}]{laion_hq}
LAION. 2024.
\newblock laion-high-resolution.

\bibitem[{Li et~al.(2023)Li, Wang, Wang, Ge, Ge, and Shan}]{li2023seed}
Li, B.; Wang, R.; Wang, G.; Ge, Y.; Ge, Y.; and Shan, Y. 2023.
\newblock Seed-bench: Benchmarking multimodal llms with generative comprehension.
\newblock \emph{arXiv preprint arXiv:2307.16125}.

\bibitem[{Li et~al.(2024)Li, Tian, Li, Deng, and He}]{mar}
Li, T.; Tian, Y.; Li, H.; Deng, M.; and He, K. 2024.
\newblock Autoregressive image generation without vector quantization.
\newblock \emph{Advances in Neural Information Processing Systems}, 37: 56424--56445.

\bibitem[{Liu et~al.(2025)Liu, Han, Xing, Yin, Wang, Cheng, Liao, Wang, Fu, Han et~al.}]{liu2025step1x}
Liu, S.; Han, Y.; Xing, P.; Yin, F.; Wang, R.; Cheng, W.; Liao, J.; Wang, Y.; Fu, H.; Han, C.; et~al. 2025.
\newblock Step1x-edit: A practical framework for general image editing.
\newblock \emph{arXiv preprint arXiv:2504.17761}.

\bibitem[{Liu et~al.(2015)Liu, Luo, Wang, and Tang}]{liu2015faceattributes}
Liu, Z.; Luo, P.; Wang, X.; and Tang, X. 2015.
\newblock Deep Learning Face Attributes in the Wild.
\newblock In \emph{Proceedings of International Conference on Computer Vision (ICCV)}.

\bibitem[{Ma et~al.(2025)Ma, Liu, Chen, Liu, Wu, Wu, Pan, Xie, Zhang, Yu et~al.}]{ma2025janusflow}
Ma, Y.; Liu, X.; Chen, X.; Liu, W.; Wu, C.; Wu, Z.; Pan, Z.; Xie, Z.; Zhang, H.; Yu, X.; et~al. 2025.
\newblock Janusflow: Harmonizing autoregression and rectified flow for unified multimodal understanding and generation.
\newblock In \emph{Proceedings of the Computer Vision and Pattern Recognition Conference}, 7739--7751.

\bibitem[{ModelScope(2025)}]{diffsynth}
ModelScope. 2025.
\newblock Diffsynth-Studio.

\bibitem[{OpenAI(2025)}]{gpt4oimagegeneration}
OpenAI. 2025.
\newblock Introducing 4o Image Generation.

\bibitem[{Pan et~al.(2023)Pan, Sun, Ge, Li, Duan, Wu, Zhang, Zhou, Qin, Wang, Dai, Qiao, and Li}]{pan2023journeydb}
Pan, J.; Sun, K.; Ge, Y.; Li, H.; Duan, H.; Wu, X.; Zhang, R.; Zhou, A.; Qin, Z.; Wang, Y.; Dai, J.; Qiao, Y.; and Li, H. 2023.
\newblock JourneyDB: A Benchmark for Generative Image Understanding.
\newblock arXiv:2307.00716.

\bibitem[{Pan et~al.(2025)Pan, Shukla, Singh, Zhao, Mishra, Wang, Xu, Chen, Li, Juefei-Xu et~al.}]{pan2025transfer}
Pan, X.; Shukla, S.~N.; Singh, A.; Zhao, Z.; Mishra, S.~K.; Wang, J.; Xu, Z.; Chen, J.; Li, K.; Juefei-Xu, F.; et~al. 2025.
\newblock Transfer between modalities with metaqueries.
\newblock \emph{arXiv preprint arXiv:2504.06256}.

\bibitem[{Podell et~al.(2024)Podell, English, Lacey, Blattmann, Dockhorn, M{\"u}ller, Penna, and Rombach}]{podell2023sdxl}
Podell, D.; English, Z.; Lacey, K.; Blattmann, A.; Dockhorn, T.; M{\"u}ller, J.; Penna, J.; and Rombach, R. 2024.
\newblock SDXL: Improving Latent Diffusion Models for High-Resolution Image Synthesis.
\newblock In \emph{ICLR}.

\bibitem[{Qu et~al.(2025)Qu, Zhang, Liu, Wang, Jiang, Gao, Ye, Du, Yuan, and Wu}]{qu2025tokenflow}
Qu, L.; Zhang, H.; Liu, Y.; Wang, X.; Jiang, Y.; Gao, Y.; Ye, H.; Du, D.~K.; Yuan, Z.; and Wu, X. 2025.
\newblock Tokenflow: Unified image tokenizer for multimodal understanding and generation.
\newblock In \emph{Proceedings of the Computer Vision and Pattern Recognition Conference}, 2545--2555.

\bibitem[{Radford et~al.(2021)Radford, Kim, Hallacy, Ramesh, Goh, Agarwal, Sastry, Askell, Mishkin, Clark et~al.}]{clip}
Radford, A.; Kim, J.~W.; Hallacy, C.; Ramesh, A.; Goh, G.; Agarwal, S.; Sastry, G.; Askell, A.; Mishkin, P.; Clark, J.; et~al. 2021.
\newblock Learning transferable visual models from natural language supervision.
\newblock In \emph{International conference on machine learning}, 8748--8763. PmLR.

\bibitem[{Raffel et~al.(2020)Raffel, Shazeer, Roberts, Lee, Narang, Matena, Zhou, Li, and Liu}]{t5}
Raffel, C.; Shazeer, N.; Roberts, A.; Lee, K.; Narang, S.; Matena, M.; Zhou, Y.; Li, W.; and Liu, P.~J. 2020.
\newblock Exploring the limits of transfer learning with a unified text-to-text transformer.
\newblock \emph{Journal of machine learning research}, 21(140): 1--67.

\bibitem[{Shi et~al.(2024)Shi, Han, Zhou, Liang, Lin, Zettlemoyer, and Yu}]{shi2024llamafusion}
Shi, W.; Han, X.; Zhou, C.; Liang, W.; Lin, X.~V.; Zettlemoyer, L.; and Yu, L. 2024.
\newblock LlamaFusion: Adapting Pretrained Language Models for Multimodal Generation.
\newblock \emph{arXiv preprint arXiv:2412.15188}.

\bibitem[{Singh et~al.(2019)Singh, Natarajan, Shah, Jiang, Chen, Batra, Parikh, and Rohrbach}]{textvqa}
Singh, A.; Natarajan, V.; Shah, M.; Jiang, Y.; Chen, X.; Batra, D.; Parikh, D.; and Rohrbach, M. 2019.
\newblock Towards vqa models that can read.
\newblock In \emph{Proceedings of the IEEE/CVF conference on computer vision and pattern recognition}, 8317--8326.

\bibitem[{Sun et~al.(2024{\natexlab{a}})Sun, Jiang, Chen, Zhang, Peng, Luo, and Yuan}]{vq}
Sun, P.; Jiang, Y.; Chen, S.; Zhang, S.; Peng, B.; Luo, P.; and Yuan, Z. 2024{\natexlab{a}}.
\newblock Autoregressive model beats diffusion: Llama for scalable image generation.
\newblock \emph{arXiv preprint arXiv:2406.06525}.

\bibitem[{Sun et~al.(2024{\natexlab{b}})Sun, Cui, Zhang, Zhang, Yu, Wang, Rao, Liu, Huang, and Wang}]{emu2}
Sun, Q.; Cui, Y.; Zhang, X.; Zhang, F.; Yu, Q.; Wang, Y.; Rao, Y.; Liu, J.; Huang, T.; and Wang, X. 2024{\natexlab{b}}.
\newblock Generative multimodal models are in-context learners.
\newblock In \emph{Proceedings of the IEEE/CVF Conference on Computer Vision and Pattern Recognition}, 14398--14409.

\bibitem[{Team(2024)}]{chameleon}
Team, C. 2024.
\newblock Chameleon: Mixed-modal early-fusion foundation models.
\newblock \emph{arXiv preprint arXiv:2405.09818}.

\bibitem[{Tong et~al.(2024{\natexlab{a}})Tong, Brown, Wu, Woo, Middepogu, Akula, Yang, Yang, Iyer, Pan et~al.}]{cambrian}
Tong, S.; Brown, E.; Wu, P.; Woo, S.; Middepogu, M.; Akula, S.~C.; Yang, J.; Yang, S.; Iyer, A.; Pan, X.; et~al. 2024{\natexlab{a}}.
\newblock Cambrian-1: A fully open, vision-centric exploration of multimodal llms.
\newblock In \emph{NeurIPS}.

\bibitem[{Tong et~al.(2024{\natexlab{b}})Tong, Fan, Zhu, Xiong, Chen, Sinha, Rabbat, LeCun, Xie, and Liu}]{metamorph}
Tong, S.; Fan, D.; Zhu, J.; Xiong, Y.; Chen, X.; Sinha, K.; Rabbat, M.; LeCun, Y.; Xie, S.; and Liu, Z. 2024{\natexlab{b}}.
\newblock MetaMorph: Multimodal Understanding and Generation via Instruction Tuning.
\newblock \emph{arXiv preprint arXiv:2412.14164}.

\bibitem[{Tuo et~al.(2023)Tuo, Xiang, He, Geng, and Xie}]{tuo2023anytext}
Tuo, Y.; Xiang, W.; He, J.-Y.; Geng, Y.; and Xie, X. 2023.
\newblock Anytext: Multilingual visual text generation and editing.
\newblock \emph{arXiv preprint arXiv:2311.03054}.

\bibitem[{Wang et~al.(2024)Wang, Zhang, Luo, Sun, Cui, Wang, Zhang, Wang, Li, Yu et~al.}]{wang2024emu3}
Wang, X.; Zhang, X.; Luo, Z.; Sun, Q.; Cui, Y.; Wang, J.; Zhang, F.; Wang, Y.; Li, Z.; Yu, Q.; et~al. 2024.
\newblock Emu3: Next-Token Prediction is All You Need.
\newblock \emph{arXiv preprint arXiv:2409.18869}.

\bibitem[{Wang et~al.(2022)Wang, Montoya, Munechika, Yang, Hoover, and Chau}]{wangDiffusionDBLargescalePrompt2022}
Wang, Z.~J.; Montoya, E.; Munechika, D.; Yang, H.; Hoover, B.; and Chau, D.~H. 2022.
\newblock {{DiffusionDB}}: {{A}} Large-Scale Prompt Gallery Dataset for Text-to-Image Generative Models.
\newblock \emph{arXiv:2210.14896 [cs]}.

\bibitem[{Wei et~al.(2024)Wei, Xiong, Ren, Du, Zhang, and Chen}]{wei2024omniedit}
Wei, C.; Xiong, Z.; Ren, W.; Du, X.; Zhang, G.; and Chen, W. 2024.
\newblock OmniEdit: Building Image Editing Generalist Models Through Specialist Supervision.
\newblock \emph{arXiv preprint arXiv:2411.07199}.

\bibitem[{Wu et~al.(2025)Wu, Chen, Wu, Ma, Liu, Pan, Liu, Xie, Yu, Ruan et~al.}]{wu2025janus}
Wu, C.; Chen, X.; Wu, Z.; Ma, Y.; Liu, X.; Pan, Z.; Liu, W.; Xie, Z.; Yu, X.; Ruan, C.; et~al. 2025.
\newblock Janus: Decoupling visual encoding for unified multimodal understanding and generation.
\newblock In \emph{Proceedings of the Computer Vision and Pattern Recognition Conference}, 12966--12977.

\bibitem[{Wu et~al.(2024)Wu, Zhang, Chen, Tang, Li, Fang, Zhu, Xie, Yin, Yi et~al.}]{wu2024vila}
Wu, Y.; Zhang, Z.; Chen, J.; Tang, H.; Li, D.; Fang, Y.; Zhu, L.; Xie, E.; Yin, H.; Yi, L.; et~al. 2024.
\newblock Vila-u: a unified foundation model integrating visual understanding and generation.
\newblock \emph{arXiv preprint arXiv:2409.04429}.

\bibitem[{XAI(2024)}]{rwqa}
XAI. 2024.
\newblock RealWorldQA.
\newblock \url{https://huggingface.co/datasets/visheratin/realworldqa}.

\bibitem[{Xiao et~al.(2025)Xiao, Wang, Zhou, Yuan, Xing, Yan, Li, Wang, Huang, and Liu}]{xiao2025omnigen}
Xiao, S.; Wang, Y.; Zhou, J.; Yuan, H.; Xing, X.; Yan, R.; Li, C.; Wang, S.; Huang, T.; and Liu, Z. 2025.
\newblock Omnigen: Unified image generation.
\newblock In \emph{Proceedings of the Computer Vision and Pattern Recognition Conference}, 13294--13304.

\bibitem[{Xie et~al.(2024)Xie, Mao, Bai, Zhang, Wang, Lin, Gu, Chen, Yang, and Shou}]{showo}
Xie, J.; Mao, W.; Bai, Z.; Zhang, D.~J.; Wang, W.; Lin, K.~Q.; Gu, Y.; Chen, Z.; Yang, Z.; and Shou, M.~Z. 2024.
\newblock Show-o: One single transformer to unify multimodal understanding and generation.
\newblock \emph{arXiv preprint arXiv:2408.12528}.

\bibitem[{Yu et~al.(2025)Yu, Chow, Yue, Pan, Wu, Wan, Li, Tang, Zhang, and Zhuang}]{yu2025anyedit}
Yu, Q.; Chow, W.; Yue, Z.; Pan, K.; Wu, Y.; Wan, X.; Li, J.; Tang, S.; Zhang, H.; and Zhuang, Y. 2025.
\newblock Anyedit: Mastering unified high-quality image editing for any idea.
\newblock In \emph{Proceedings of the Computer Vision and Pattern Recognition Conference}, 26125--26135.

\bibitem[{Yue et~al.(2024)Yue, Ni, Zhang, Zheng, Liu, Zhang, Stevens, Jiang, Ren, Sun, Wei, Yu, Yuan, Sun, Yin, Zheng, Yang, Liu, Huang, Sun, Su, and Chen}]{yue2023mmmu}
Yue, X.; Ni, Y.; Zhang, K.; Zheng, T.; Liu, R.; Zhang, G.; Stevens, S.; Jiang, D.; Ren, W.; Sun, Y.; Wei, C.; Yu, B.; Yuan, R.; Sun, R.; Yin, M.; Zheng, B.; Yang, Z.; Liu, Y.; Huang, W.; Sun, H.; Su, Y.; and Chen, W. 2024.
\newblock MMMU: A Massive Multi-discipline Multimodal Understanding and Reasoning Benchmark for Expert AGI.
\newblock In \emph{Proceedings of CVPR}.

\bibitem[{Zhai et~al.(2023)Zhai, Mustafa, Kolesnikov, and Beyer}]{siglip}
Zhai, X.; Mustafa, B.; Kolesnikov, A.; and Beyer, L. 2023.
\newblock Sigmoid loss for language image pre-training.
\newblock In \emph{Proceedings of the IEEE/CVF international conference on computer vision}, 11975--11986.

\bibitem[{Zhang et~al.(2025)Zhang, Duan, Wang, Chen, and Zhang}]{zhang2025eligen}
Zhang, H.; Duan, Z.; Wang, X.; Chen, Y.; and Zhang, Y. 2025.
\newblock EliGen: Entity-Level Controlled Image Generation with Regional Attention.
\newblock \emph{arXiv preprint arXiv:2501.01097}.

\bibitem[{Zhang et~al.(2023)Zhang, Mo, Chen, Sun, and Su}]{zhang2023magicbrush}
Zhang, K.; Mo, L.; Chen, W.; Sun, H.; and Su, Y. 2023.
\newblock Magicbrush: A manually annotated dataset for instruction-guided image editing.
\newblock \emph{Advances in Neural Information Processing Systems}, 36: 31428--31449.

\bibitem[{Zhao et~al.(2024{\natexlab{a}})Zhao, Ma, Chen, Si, Wu, An, Yu, Zhang, Li, and Chang}]{zhao2024ultraedit}
Zhao, H.; Ma, X.~S.; Chen, L.; Si, S.; Wu, R.; An, K.; Yu, P.; Zhang, M.; Li, Q.; and Chang, B. 2024{\natexlab{a}}.
\newblock Ultraedit: Instruction-based fine-grained image editing at scale.
\newblock \emph{Advances in Neural Information Processing Systems}, 37: 3058--3093.

\bibitem[{Zhao et~al.(2024{\natexlab{b}})Zhao, Huang, Hu, Wang, Mao, Zhang, Jiang, Wu, Ai, Wang, Zhou, and Chen}]{zhao2024swiftascalablelightweightinfrastructure}
Zhao, Y.; Huang, J.; Hu, J.; Wang, X.; Mao, Y.; Zhang, D.; Jiang, Z.; Wu, Z.; Ai, B.; Wang, A.; Zhou, W.; and Chen, Y. 2024{\natexlab{b}}.
\newblock SWIFT:A Scalable lightWeight Infrastructure for Fine-Tuning.
\newblock arXiv:2408.05517.

\bibitem[{Zhou et~al.(2025)Zhou, Yu, Babu, Tirumala, Yasunaga, Shamis, Kahn, Ma, Zettlemoyer, and Levy}]{transfusion}
Zhou, C.; Yu, L.; Babu, A.; Tirumala, K.; Yasunaga, M.; Shamis, L.; Kahn, J.; Ma, X.; Zettlemoyer, L.; and Levy, O. 2025.
\newblock Transfusion: Predict the next token and diffuse images with one multi-modal model.
\newblock In \emph{ICLR}.

\end{thebibliography}
\appendix
\clearpage
\section{Dataset Construction Details}

\subsection{Dataset distribution}
\begin{figure}[thp] 
    \centering
    \includegraphics[width=1.0\linewidth]{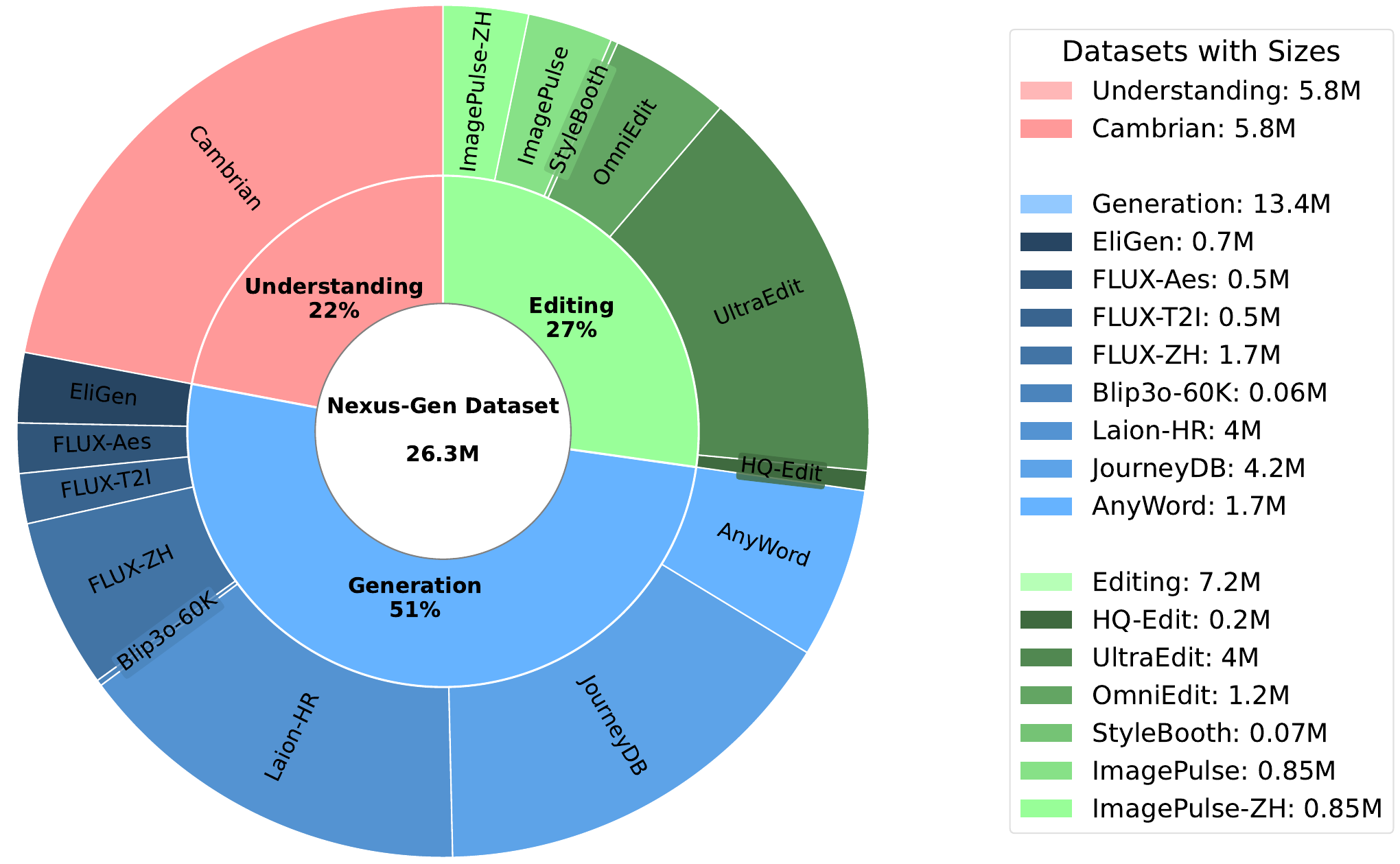}
    \caption{Dataset distribution of our Nexus-Gen dataset.}
    \label{fig:dataset_distribution}
\end{figure}

We construct a unified dataset covering image understanding, generation and editing tasks with 26.3 million samples. Detailed dataset distribution is presented in Figure~\ref{fig:dataset_distribution}.

\paragraph{Image Understanding}
This task is structured with multimodal inputs (image-text pairs) and text-only outputs, which serves as a direct indicator of model’s chat and understanding ability. While MLLMs inherently possess such cross-modal reasoning capabilities, this task is still critical during training  to prevent capacity degradation. We adopt Cambrian-7M \cite{cambrian} as the data source, a comprehensive dataset spanning multiple domains including optical character recognition, general visual question answering, language, counting, code, math and science tasks. To improve data quality, we re-annotate the answers for all samples with Qwen2.5-VL-72B \cite{bai2025qwen2.5-vl}.

\paragraph{Image Generation}
The input for this task is the textual description, and the output is an image. Our data sources comprise Journey DB \cite{pan2023journeydb}, AnyWord \cite{tuo2023anytext}, Laion-High-Resolution \cite{laion_hq}, EliGen TrainSet \cite{zhang2025eligen}, FLUX-Aes \cite{flux_aes}, FLUX-T2I \cite{flux_T2I} and Blip3o-60K \cite{chen2025blip3}. To enhance annotation diversity, we employ a dual-captioning paradigm via Qwen2.5-VL-72B, generating both concise captions and elaborate descriptions for each image. During training, we stochastically sample these annotations with stratified ratios (20\% concise vs. 80\% elaborate) to balance brevity and contextual granularity.

\paragraph{Image Editing}
The input for editing task consists of an image and its corresponding editing instruction, and the output denotes the edited image. Our data sources encompass datasets such as HQ-Edit \cite{hqedit}, UltraEdit \cite{zhao2024ultraedit}, OmniEdit \cite{wei2024omniedit}, and StyleBooth \cite{han2024stylebooth}. These datasets cover an extensive range of image editing operations, including object-level manipulations, color adjustments, and style transformations. However, these datasets exhibit notable limitations in aesthetic quality. However, they exhibit notable limitations in aesthetic quality and visual fidelity. To address this, we construct the high-quality ImagePulse dataset and integrate it into our training corpus.

\paragraph{Bilingual Annotations}
Apart from the Chinese samples present in image understanding datasets, all aforementioned datasets are annotated exclusively in English. To endow Nexus-Gen with bilingual (Chinese-English) capabilities for both image generation and editing, additional Chinese image generation and editing samples are incorporated into the dataset. To this end, we perform Chinese re-annotation on several generation subsets (namely EliGen, FLUX-Aes, and FLUX-T2I) as well as the ImagePulse editing dataset. This process yielded a corpus of 2.5 million Chinese training samples, comprising the FLUX-ZH and ImagePulse-ZH subsets illustrated in Figure~\ref{fig:dataset_distribution}.

\begin{table*}[t]
\centering
\begin{tabular}{lcccc}
\toprule
Training Phase     & \multicolumn{2}{c}{Multi-task Pretraining for Autoregressive Model} & \multicolumn{2}{c}{Conditional Adaption for Vision Decoder}          \\ \midrule
Training Target     & Large Scale Pretraining     & Aesthetic Fine-tuning     & Generation Decoder & Editing Decoder \\ \midrule
Learning Rate      & 1e-5    & 1e-5    & 1e-5               & 1e-5            \\
LR Scheduler       & Cosine  & Cosine  & Constant           & Constant        \\
Batch Size         & 512     & 512     & 128                & 128             \\
Total Steps        & 100 K   & 10 K    & 20 K               & 10 K            \\
Warm-up Steps      & 7500    & 800     & 100                & 100             \\
Total Samples (Million) & 26 & 4       & 2                  & 1           \\
Data Ratio (Und:Gen:Edit) & 1:2:1   & 1:2:1   & 0:1:0              & 0:0:1           \\
\bottomrule
\end{tabular}
\caption{Detailed hyperparameters for training Nexus-Gen. Data ratio refers to the ratio of image understanding data, generation data and editing data.}
\label{tab:hyperparam}
\end{table*}

\begin{figure*}[tp] 
    \centering
    \includegraphics[width=0.9\linewidth]{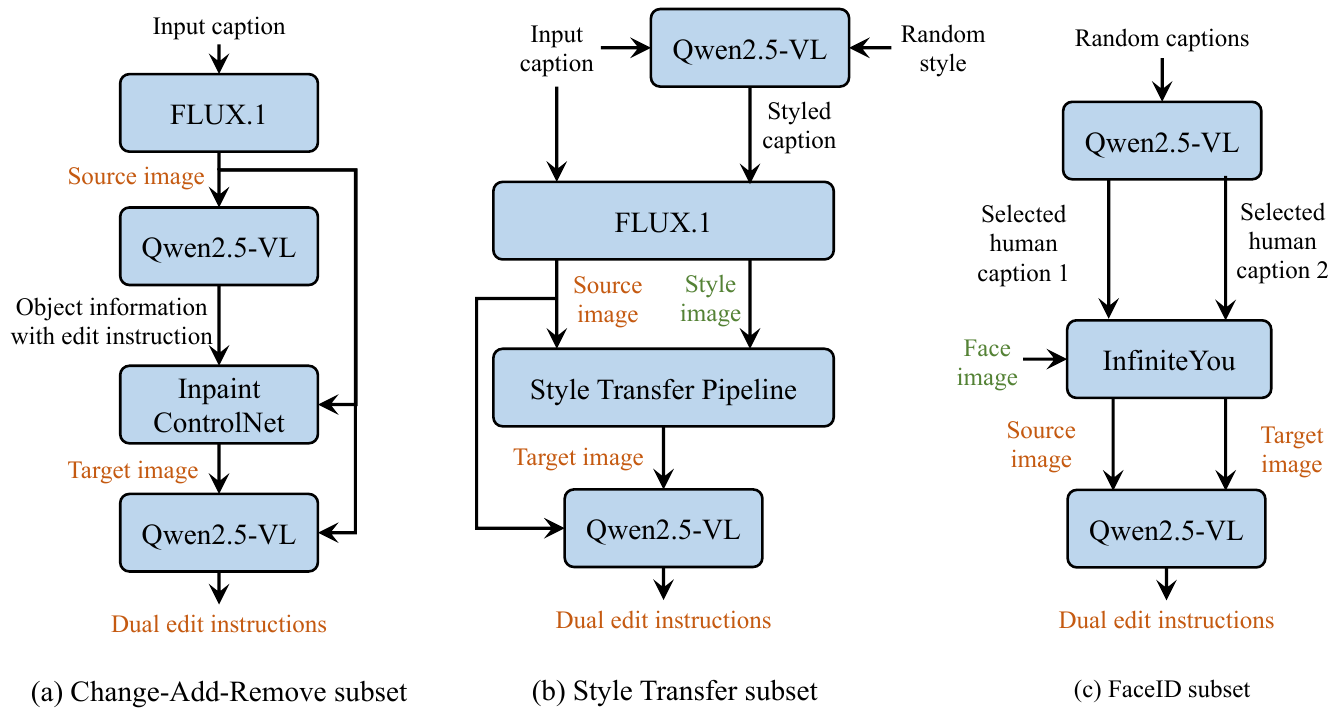}
    \caption{The dataset construction pipeline for three subsets of ImagePulse.}
    \label{fig:imagepulse}
\end{figure*}

\subsection{Construction Pipeline for ImagePulse}
Each sample in ImagePulse contains: (1) pristine image pairs synthesized by FLUX.1-Dev \cite{flux}, and (2) semantically-rich editing instruction generated by Qwen2.5-VL-72B \cite{bai2025qwen2.5-vl}. For diverse editing tasks, the dataset is partitioned into three specialized subsets: Change-Add-Remove, Style Transfer, and FaceID. The workflow for each subset is illustrated in Figure~\ref{fig:imagepulse}
\paragraph{Change-Add-Remove}
This subset focuses on object-level image manipulations, including modifying object attributes (such as shape, material, and color), adding objects, and removing objects. The dataset construction pipeline is illustrated in Figure~\ref{fig:imagepulse}(a). First, we randomly sample a caption from the DiffusionDB \cite{wangDiffusionDBLargescalePrompt2022} dataset and synthesize a source image using the FLUX.1-Dev model. Next, the Qwen2.5-VL model extracts object information, comprising semantic descriptions and spatial locations, from this image and generates corresponding editing instructions. Subsequently, we use the extracted spatial locations and editing instructions to modify specified regions via Inpaint ControlNet \cite{FLUX-Controlnet}, producing the target image. Finally, to maximize data utility, Qwen2.5-VL generates bidirectional editing instructions between the source and target images.

\paragraph{Style Transfer}
This subset tackles the problem of image style transfer, with its construction process illustrated in Figure~\ref{fig:imagepulse}(b). First, a randomly selected input caption and a target style prompt are processed by Qwen2.5-VL to generate a style-transformed caption. Subsequently, the original input caption and the generated style-transformed caption are leveraged to synthesize the source image and the style image, respectively. Crucially, the style image exhibits significant structural divergence from the source image, rendering it unsuitable as the target image. To derive the target image, we devise a Style Transfer Pipeline integrating ControlNet, SDXL, InstantStyle, and IP-Adapter models. This pipeline effectively fuses the structural framework of the source image with the stylistic properties of the style image. Finally, we generate corresponding dual editing instructions.

\paragraph{FaceID}
The unified architecture of Nexus-Gen facilitates tackling particularly challenging conceptual editing tasks by harnessing advanced image generation capabilities. To validate this capability, we construct the FaceID subset, which focuses on executing high-variation conceptual edits while preserving subject identity. The dataset construction process is illustrated in Figure~\ref{fig:imagepulse}(c). Initially, Qwen2.5-VL selects two captions containing human descriptions from a pool of randomly sampled image captions. Subsequently, a facial image is randomly sampled from the CelebA-HQ-Face \cite{liu2015faceattributes} dataset. Using these two captions and the facial image, we synthesize the source and target images via the InfiniteYou \cite{jiang2025infiniteyou} model. These images exhibit identical subject identity while differing significantly in pose and scene context. Finally, dual editing instructions corresponding to the source and target images are generated by Qwen2.5-VL.


\section{Implementation Details}
Nexus-Gen employs Qwen2.5-VL as its autoregressive model and FLUX.1-Dev as the generation and editing decoder, utilizing a multi-stage training strategy to optimize each component separately. The autoregressive model undergoes training within the ms-swift \cite{zhao2024swiftascalablelightweightinfrastructure} framework, comprising pretraining followed by an aesthetic fine-tuning stage. The generation and editing decoders are trained using Diffsynth-Studio \cite{diffsynth}. Detailed training hyperparameters are provided in Table~\ref{tab:hyperparam}. During inference, the generation decoder utilizes classifier-free guidance with a scale of 3.0.

\section{More Qualitative Results}
In this section, we present additional qualitative results for Nexus-Gen, focusing on image generation and editing tasks.
\subsection{Image Generation}
\begin{figure*}[tp] 
    \centering
    \includegraphics[width=1.0\linewidth]{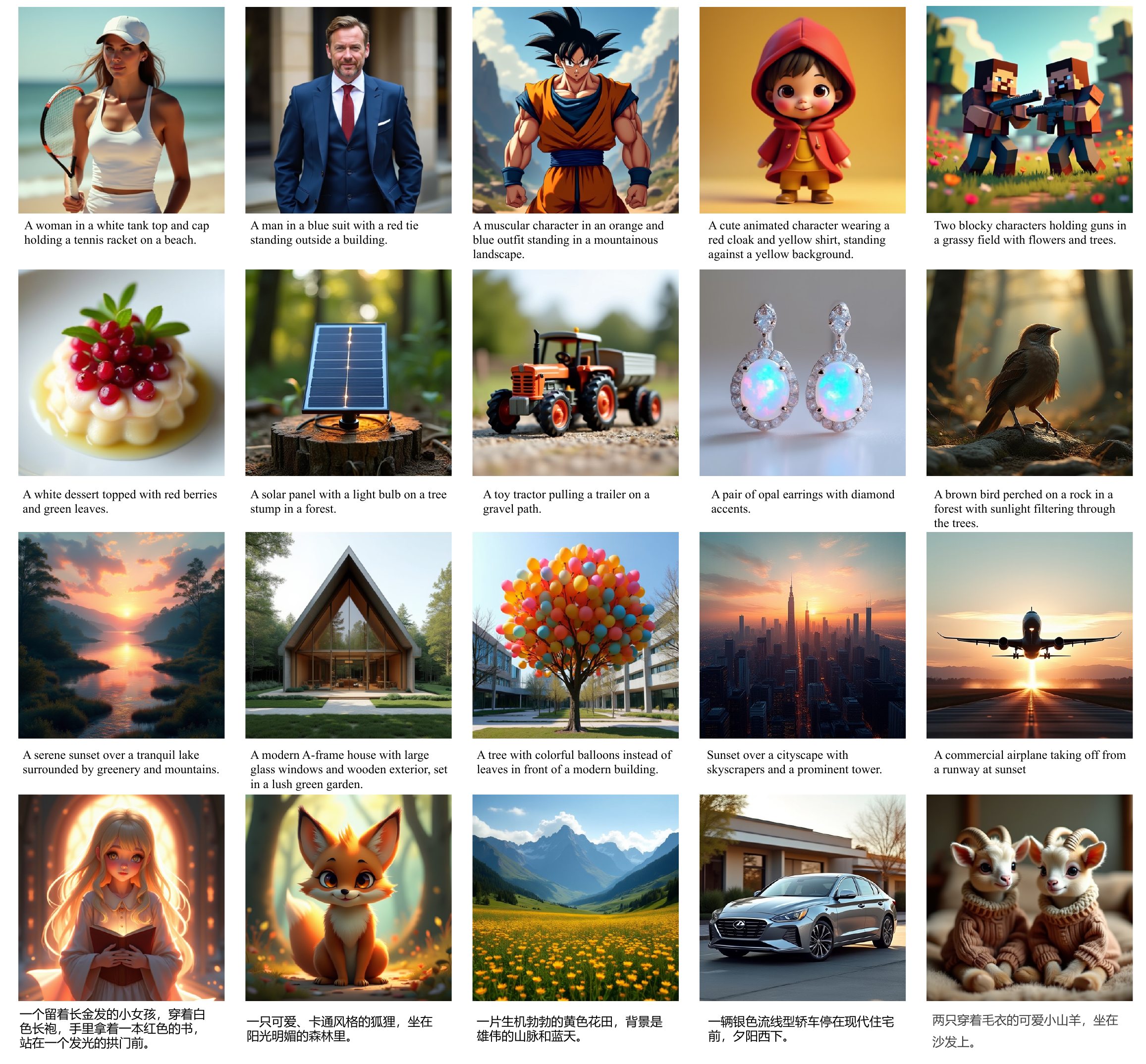}
    \caption{Qualitative image generation results of Nexus-Gen.}
    \label{fig:generation}
\end{figure*}

Figure~\ref{fig:generation} showcases representative high-fidelity images synthesized by Nexus-Gen, demonstrating the model's capability to accurately interpret semantic information from textual descriptions and translate them into visually coherent outputs. Owing to the incorporation of bilingual image generation datasets, Nexus-Gen is capable of processing inputs and generating outputs in both English and Chinese.
\subsection{Image Editing}
\begin{figure*}[tp] 
    \centering
    \includegraphics[width=1.0\linewidth]{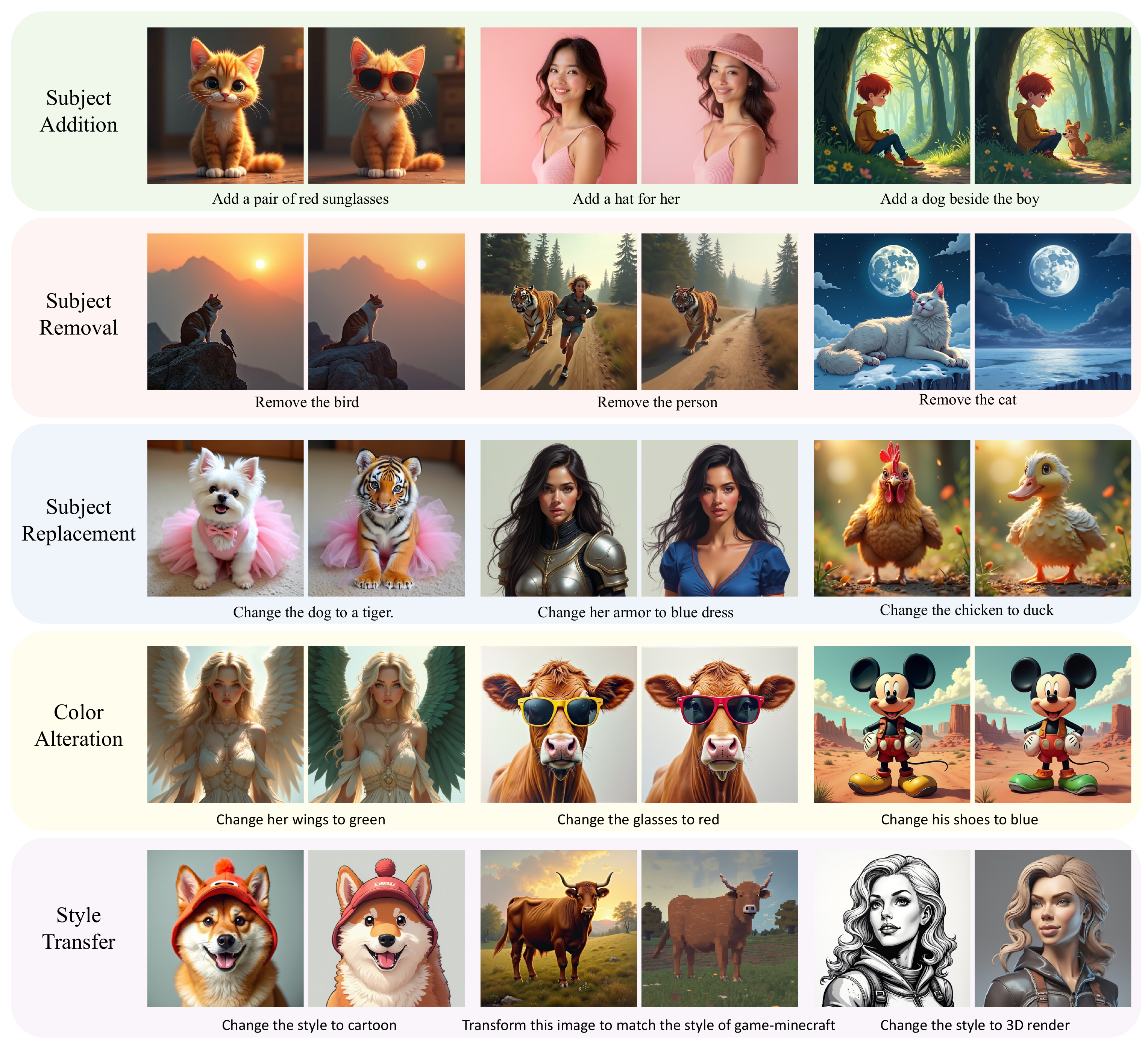}
    \caption{Qualitative image editing results of Nexus-Gen.}
    \label{fig:editing}
\end{figure*}

The image editing capabilities of Nexus-Gen are demonstrated in Figure~\ref{fig:editing}, which exhibits seamless handling of diverse workflows including subject addition, removal, replacement, color alteration, and style transfer. It can be observed that Nexus-Gen demonstrates remarkable proficiency in both preserving non-edited regions and executing editing instructions.

\section{Limitations}
Despite its capabilities in image understanding, generation, and editing, Nexus-Gen exhibits distinct limitations. First, with a total training dataset of 26 million samples, its scale remains substantially smaller than either specialized single-task models or unified counterparts trained on hyper-scale datasets. Consequently, the model may exhibit sensitivity to image generation prompts and often requires specific instruction templates for optimal editing performance. Second, unlike VAE latent spaces that enable precise pixel-level reconstruction, Nexus-Gen's unified image space operates primarily at the semantic feature level, resulting in inherent reconstruction fidelity limitations. Third, the visual reasoning capabilities of Nexus-Gen remains unexplored.

\end{document}